\PassOptionsToPackage{varqu}{inconsolata}  
\documentclass{article} 
\usepackage[final]{colm2026_conference}

\usepackage{hyperref}
\usepackage{microtype}
\usepackage{url}
\usepackage{booktabs}
\usepackage{graphicx}
\usepackage{dirtytalk}
\usepackage{amssymb}
\usepackage{amsmath}
\usepackage[capitalise,noabbrev]{cleveref}
\usepackage{listings}
\usepackage{xcolor}
\usepackage{caption}
\usepackage{subcaption}
\usepackage{adjustbox}
\usepackage{float}
\usepackage{multirow}

\usepackage{algorithm}
\usepackage{algpseudocode}

\usepackage[T1]{fontenc}
\usepackage{inconsolata}

="[fonts/NotoSerifSC-VariableFont_wght.ttf]" at 10pt

="[fonts/NotoSerifDevanagari-Variable.ttf]" at 10pt

="[fonts/NotoSerif-VariableFont_wdth,wght.ttf]" at 10pt

\newcommand{\cyr}[1]{{\cyrfont #1}}

\newcommand{\sq}{\textquotesingle}

\crefname{equation}{Eq}{Eqs}
\Crefname{equation}{Eq}{Eqs}

\definecolor{darkblue}{rgb}{0, 0, 0.5}
\definecolor{stringcolor}{RGB}{186,33,33}  
\definecolor{highlightbg}{RGB}{255,255,180}  

\hypersetup{colorlinks=true, citecolor=darkblue, linkcolor=darkblue, urlcolor=darkblue}

\DeclareCaptionType{listing}[Listing][List of Listings]

\lstdefinestyle{numberedlist} {
    language=Python,
    basicstyle=\ttfamily\small,
    keepspaces=true,
    columns=flexible,
    numbers=left,
    numberstyle=\tiny\color{gray},
    numbersep=5pt,
    stepnumber=1,
    stringstyle=\color{stringcolor},
    showstringspaces=false,
    escapeinside={@}{@},
}

\lstdefinestyle{smallsize} {
    language=Python,
    basicstyle=\ttfamily\small,
    keepspaces=true,
    columns=flexible,
    stringstyle=\color{stringcolor},
    showstringspaces=false,
    escapeinside={@}{@},
}

\title{Faster Superword Tokenization}

\author{Craig W. Schmidt \& Chris Tanner\thanks{Chris Tanner is also affiliated with MIT in Cambridge, MA, USA.} \\
Kensho Technologies\\
Cambridge, MA 02138, USA \\
\texttt{\{craig.schmidt,chris.tanner\}@kensho.com} \\
\And
Yuval Pinter \\
Institute for Applied AI Research \\
Ben-Gurion University of the Negev \\
Beer Sheva 8410501, Israel \\
\texttt{uvp@cs.bgu.ac.il} \\
}

\begin{document}

\ifcolmsubmission
\linenumbers
\fi

\maketitle

\begin{abstract}
Byte Pair Encoding (BPE) is a widely used tokenization algorithm, whose tokens cannot extend across pre-tokenization boundaries, functionally limiting it to representing at most full words. The BoundlessBPE and SuperBPE algorithms extend and improve BPE by relaxing this limitation and allowing the formation of superwords, which are combinations of pretokens that form phrases. However, previous implementations were impractical to train: for example, BoundlessBPE took 4.7 CPU days to train on 1GB of data. We show that supermerge candidates (i.e., two or more consecutive pretokens eligible to form a supermerge) can be aggregated by frequency much like pretokens. This avoids keeping full documents in memory, as the original implementations of BoundlessBPE and SuperBPE required, leading to a significant training speedup. We present a two-phase formulation of BoundlessBPE that separates first-phase learning of ordinary merges from second-phase learning of supermerges, producing identical results to the original implementation. We also show a near-equivalence in resulting tokens between two-phase BoundlessBPE and SuperBPE, with the difference being that a manually selected hyperparameter used in SuperBPE can be automatically determined in the second phase of BoundlessBPE. These changes enable a much faster implementation, allowing training on the same 1GB of data in 603 and 593 seconds for BoundlessBPE and SuperBPE, respectively, a more than 600-fold increase in speed. For each of BoundlessBPE, SuperBPE, and BPE, we open-source both a reference Python implementation and a fast Rust implementation.
\end{abstract}

\section{Introduction}

Byte Pair Encoding \citep[BPE;][]{sennrich-etal-2016-improving,gage1994new} is a subword tokenization method used by virtually all current large language models.
Before its core algorithmic process, BPE first splits text with a regular expression (regex) into \textit{pretokens}, which are segments that roughly represent distinct words, punctuation, whitespace, or numbers.
Pretoken boundaries, which cannot be crossed when forming tokens, act as a useful inductive bias that avoids linguistically implausible tokens.
Because boundaries cannot be crossed, each pretoken is tokenized independently.
The vocabulary is created through a simple iterative procedure where pretokens start as sequences of individual byte tokens, and at each step the most frequent adjacent pair of tokens in the corpus is merged into a single combined token.
This new token is added to the vocabulary and replaces its constituent tokens in the entire corpus.
This process repeats until the vocabulary reaches a target size.
The result is an ordered list of merge rules, which are later re-applied in the same order to tokenize new text during inference.

Following a BPE training run, 75--90\% of pretokens are tokenized as a single token, leaving little compression to be gained within pretoken boundaries~\citep{reddy2025enoughdiminishingreturnstokenization}. At the same time, many low-frequency word tokens occupy vocabulary slots that would be better spent on higher-frequency multiword phrases, which the pretoken constraint forbids.
BoundlessBPE~\citep{schmidt2025boundlessbytepairencoding} and SuperBPE~\citep{liu2025superbpespacetravellanguage}, developed concurrently and independently, relax this constraint by introducing \textit{superwords}.
A superword is formed when two adjacent pretokens, each represented as a single token, are merged into a combined token through a \textit{supermerge}.
For example, \texttt{\sq ~to\sq} and \texttt{\sq ~be\sq} might be combined to form the superword \texttt{\sq ~to~be\sq}.
BoundlessBPE outperforms BPE on compression and R\'enyi efficiency~\citep{zouhar-etal-2023-tokenization}, producing a more uniform token distribution over a training corpus.
SuperBPE also reported an average improvement of 4.0\% in accuracy over a BPE baseline across 30 downstream tasks. 

One drawback of both BoundlessBPE and SuperBPE is training speed.
BoundlessBPE reported taking 4.7 CPU days to train on only 1GB of data (a fraction of a typical training corpus), compared to 59s for the Hugging Face BPE implementation\footnote{{\scriptsize\url{https://github.com/huggingface/tokenizers}}} on the same data.
SuperBPE trained on 10GB of data in \say{a few hours} using parallel code on 100 CPUs.
This difference in training time stems largely from pretoken aggregation.
With BPE, tokenization is prohibited from creating tokens that span across pre-tokenization boundaries, so the training can work solely with an \emph{aggregated} form of the corpus: a table of the unique pretokens, each paired with the number of times it occurs.
Thus, even though the pretoken \texttt{\sq ~the\sq} may appear millions of times in the training data, it is stored once, and any merge within it is weighted by that occurrence count rather than being recomputed for each copy.
This gives BPE exactly the same merge statistics as scanning every occurrence, at a fraction of the work, and has been present in BPE since the original implementation by \citet{sennrich-etal-2016-improving}.\footnote{{\scriptsize\url{https://github.com/rsennrich/subword-nmt/blob/92d61/subword_nmt/learn_bpe.py\#L80}}}
In contrast, the implementations of supermerges presented in BoundlessBPE and SuperBPE keep a copy of all pretokens for each document in memory to provide context for supermerges (\citet{schmidt2025boundlessbytepairencoding}, Appendix B; \citet{liu2025superbpespacetravellanguage}, Section 2.2).
Since the total time to perform a merge scales linearly with the number of \emph{unique} pretokens in which the pair appears, more aggregation---representing many occurrences with fewer unique entries---directly leads to faster training.

In this work, we present a method of pretoken aggregation compatible with BoundlessBPE and SuperBPE, reducing the 4.7 CPU days to 603 and 593 CPU seconds respectively (improving the computational time of the original algorithms by over 600-fold).\footnote{Following \citet{schmidt2025boundlessbytepairencoding}, these times are for 1GB of data, roughly one-sixth of the full MiniPile corpus of 1M documents. \cref{sec:computationalresults} reports scaling up to the 5.9GB corpus.}
To enable this, we present a two-phase BoundlessBPE algorithm that gives identical results to the original method.
In the first phase, an ordinary BPE tokenizer is trained to produce the entire desired vocabulary size.
In the second phase, the resulting merge rules are compared against the best available supermerges, giving a final mix of supermerges and a subset of the merges from the first phase.
This two-phase structure aligns BoundlessBPE with SuperBPE, which already operates in two phases, enabling a unified aggregation approach for both.
For scripts that do not use space to delimit words, we also introduce a script-specific pre-tokenization strategy that treats each multi-byte character as its own pretoken.
This prevents the formation of tokens composed of partial bytes from two or more characters---a known source of tokens that are not valid UTF-8~\citep{land2025bpestaysscriptstructured,jang-etal-2025-improbable,firestone2025utf8plumbingbyteleveltokenizers}.
Supermerges can then assemble the complete characters back into phrases as counts permit (\cref{app:pretokenization}).
We also introduce an optional step that breaks longer supermerge candidates into shorter ones with higher aggregate counts (\cref{app:nonspacedelim}).

One major contribution of this work is an open-source library\footnote{\scriptsize{\url{https://github.com/kensho-technologies/fastboundlessbpe}}} implementing BoundlessBPE and SuperBPE that can be trained on larger training datasets in a reasonable time.  We provide two implementations that give identical results: a reference Python implementation for easy experimentation, and a faster Rust version for practitioners interested only in using the tokenizer.\footnote{The first-phase BPE tokenizer can also be used on its own. As with the original BoundlessBPE implementation, this code supports PickyBPE deletions from \citet{chizhov-etal-2024-bpe}.  However, all results in this paper were run without any PickyBPE deletions.} 
The library can also export a trained model to the Hugging Face and \texttt{tiktoken} formats for use in standard pipelines, approximating a BoundlessBPE or SuperBPE model with single-pass BPE (\cref{app:export}).

Every mention of BoundlessBPE and SuperBPE in this paper refers to our fast versions of the algorithms from \citet{schmidt2025boundlessbytepairencoding}\footnote{\scriptsize{\url{https://github.com/kensho-technologies/boundlessbpe}}} and \citet{liu2025superbpespacetravellanguage},\footnote{\scriptsize{\url{https://superbpe.github.io/}}} rather than the original implementations. In particular, our SuperBPE second phase restricts merges to pairs of whole single-token pretokens, matching the supermerge eligibility of BoundlessBPE. In contrast, the original SuperBPE relaxes pre-tokenization entirely in its second stage, allowing it to merge arbitrary adjacent token pairs rather than just whole pretokens. The variant we use allows us to consider the two methods within one framework, and it makes the equivalence of \cref{sec:relationship} exact.

\section{Aggregation Procedure}
\label{sec:aggregation}

\begin{figure}[!ht]
    \centering
    \includegraphics[width=\linewidth]{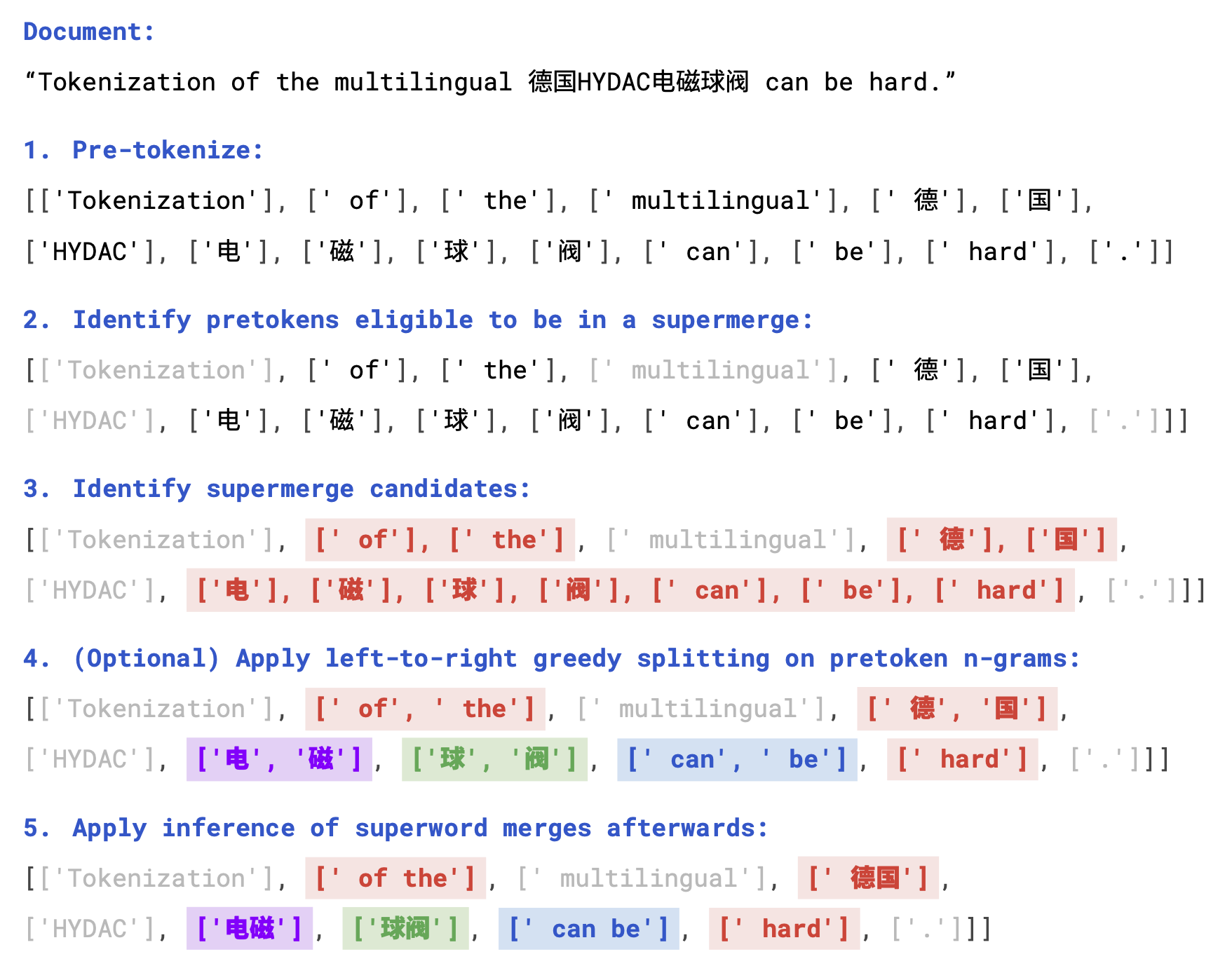}
    \caption{Illustrative example of the aggregation process using a hand-constructed vocabulary. Light gray indicates pretokens that are either composed of multiple tokens or do not include a letter, rendering them ineligible for supermerges. Red indicates supermerge candidates, and purple, green, and blue indicate greedy splits of some candidates.}
    \label{fig:example}
\end{figure}

Because its tokens cannot extend beyond pretoken boundaries, BPE only needs to store pretokens aggregated by count, giving it a low memory footprint. Learning supermerges, in contrast, requires maintaining the context of potential superwords since they can span across pretoken boundaries. However, documents need not be stored in memory in their entirety, as in the initial implementations of BoundlessBPE and SuperBPE; instead, they can be decomposed into individual supermerge candidates which are aggregated by count. \cref{fig:example} gives an illustrative example of the five steps in this aggregation process.

In step 1, we apply pre-tokenization to the document.
Pre-tokenization is the process of breaking a document into pretokens, which are segments roughly representing distinct words, punctuation, whitespace, or numbers.
This split is usually accomplished using a regular expression (regex).\footnote{We use the GPT-4o regex described in \cref{app:pretokenization}.}
No token can cross a pretoken boundary, which is a useful inductive bias.
In particular it limits spaces to only appear as the first character of a word, or to be part of a run of whitespace.
Our implementation also optionally splits the non-space-delimited scripts of Chinese, Japanese, Thai, Burmese, Khmer, and Lao into single characters, which can later be recombined into phrases using superwords (see \cref{app:pretokenization} for more details).
The resulting pretokens are the ones aggregated for the ordinary BPE training run, but not all of them will qualify for supermerges.

In step 2, after running the ordinary BPE algorithm and obtaining a vocabulary and merge list, we identify the pretokens that are eligible to be included in a supermerge.
To be eligible, a pretoken must contain at least one Unicode letter (matching the regex \texttt{\textbackslash p\{L\}}) and be represented by exactly one token.
The Unicode letter requirement avoids merging pretokens that consist solely of numbers or punctuation (e.g., \texttt{[\sq .\sq ]}), which pre-tokenization typically keeps separate.\footnote{Any regex can be used here, but one will usually want to restrict supermerges in some form.}
A pretoken must also be tokenized as a single token in order to participate in a supermerge, meaning that it appears in the ordinary BPE's resulting vocabulary.
In the example, this step rules out \texttt{[\sq Tokenization\sq ]}, \texttt{[\sq ~multilingual\sq ]}, and \texttt{[\sq HYDAC\sq ]} from participating in a supermerge because they are split into two tokens.
Ineligible pretokens are shown in light gray.

In step 3, we identify \textit{supermerge candidates}, which are two or more consecutive pretokens eligible to form a supermerge.
Supermerge candidates are shown in red.\footnote{Note that we enforce that pretokens contain only a single script, but we allow supermerges across scripts. In practice, the low counts for cross-script merges should largely prevent them from entering the vocabulary, though this can also be imposed as a hard constraint.}
A run of eligible pretokens is broken by any intervening ineligible (gray) pretoken from step 2, since an ineligible pretoken cannot join a supermerge.
The supermerge candidates from this stage (or, if step 4 is included, from that stage) are then aggregated by frequency as input to the second phase of BoundlessBPE.

Step 4 is an optional step where the supermerge candidates found in step 3 are split into smaller chunks based on pretoken $n$-gram frequency, leading to increased aggregation and thus improved training speed.
In the example, {\cjkfont 德国} translates as \say{German}, {\cjkfont 电磁} is \say{electromagnetic} or \say{solenoid},  and {\cjkfont 球阀} is \say{ball valve}.
\cref{app:nonspacedelim} describes how we compute common $n$-grams and break these longer runs into shorter segments using a left-to-right greedy heuristic.
In this example, it might break {\cjkfont 电磁球阀} into the two more common two-character phrases, shown in purple and green.
It might also split \texttt{[\sq ~can\sq ,\sq ~be\sq ]} from \texttt{[\sq ~hard\sq ]}.
We have not used this optional step in our computational results on English text, and defer further discussion to \cref{app:nonspacedelim}.

Step 5 shows a hypothetical example of what might happen at inference time, after supermerges are applied.
A supermerge combines adjacent single-token pretokens, regardless of script.
For space-delimited scripts these pretokens are whole words, so it joins them into a multi-word phrase, for example \texttt{[\sq ~of\sq ,\sq ~the\sq]} into \texttt{[\sq ~of~the\sq]}.
For the non-space-delimited scripts that are split character-by-character in step 1, the same operation instead reassembles single characters into a multi-character word such as {\cjkfont 电磁}.
The mechanism is identical in both cases; only what counts as a pretoken differs.
See \cref{sec:inference} for more about the inference procedure.

\begin{figure}[!ht]
    \centering
    \includegraphics[width=0.6\linewidth]{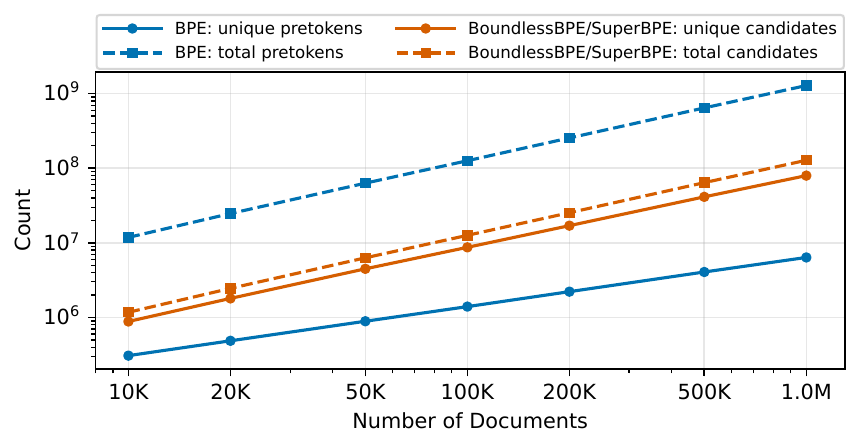}
    \caption{Unique and total counts versus the number of MiniPile documents, for BPE pretokens and for BoundlessBPE/SuperBPE supermerge candidates.}
    \label{fig:pretokenchunks}
\end{figure}

Aggregating the supermerge candidates yields a more compact representation of the training data for input to the superword tokenization phase.
The time to perform a supermerge is linear with respect to the number of supermerge candidates that contain it, improving training time in practical settings.
\cref{fig:pretokenchunks} shows the number of pretokens and supermerge candidates for incremental fractions of the 1M MiniPile documents \citep{kaddour2023minipilechallengedataefficientlanguage}.\footnote{See {\scriptsize \url{https://huggingface.co/datasets/JeanKaddour/minipile}}. MiniPile was used for training throughout this paper, with the exception of \cref{app:nonspacedelim}. It consists of 1M documents and is 5.9GB
in size.}
When aggregating pretokens for ordinary BPE, the ratio of unique entries to total occurrences (the \textit{type-token ratio}) is very low, so relatively few unique pretokens represent the entire corpus.
There are fewer total supermerge candidates than pretokens, since each candidate spans two or more pretokens and ineligible pretokens are skipped, but they have less redundancy, so aggregation provides a smaller benefit.

\section{Two-Phase BoundlessBPE}

\begin{figure}[!ht]
    \centering
    \includegraphics[width=0.95\linewidth]{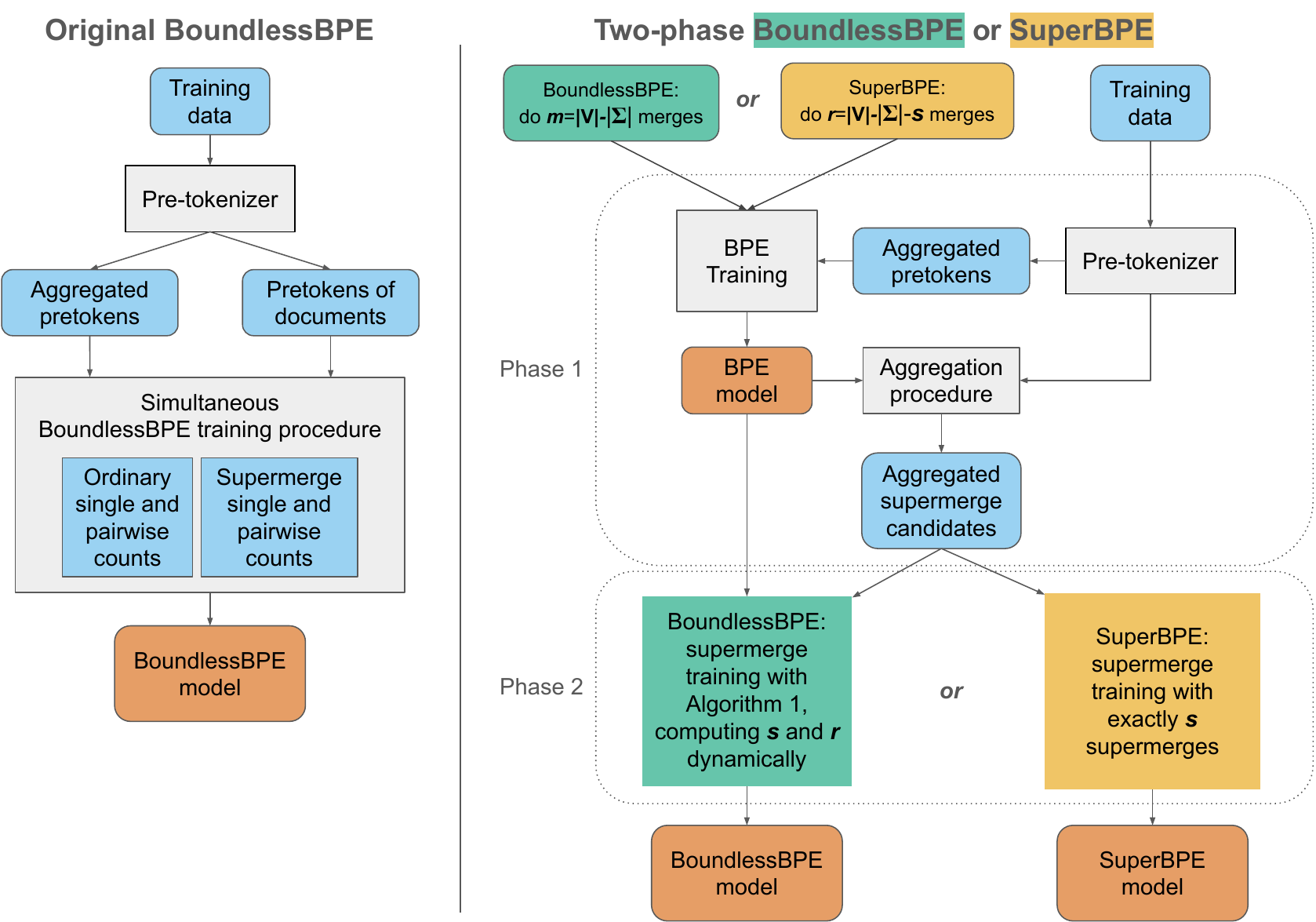}
    \caption{Comparison of original BoundlessBPE, two-phase BoundlessBPE, and SuperBPE. The original BoundlessBPE (left) performs ordinary merges and supermerges simultaneously. The two-phase approach (right) decomposes this into stages. In Phase~1, the input is pre-tokenized, a standard BPE model is trained, and supermerge candidates are identified via an aggregation procedure (\cref{sec:aggregation}). BoundlessBPE and SuperBPE follow the same Phase~1 process but differ in the number of BPE merges performed: BoundlessBPE trains to the full vocabulary size, while SuperBPE reserves $s$ merges for supermerges. In Phase~2, BoundlessBPE uses \cref{alg:secondphase} to dynamically determine how many merges are ordinary versus supermerges, while SuperBPE selects a fixed number of $s$ supermerges.}
    \label{fig:flowchart}
\end{figure}

In this section we introduce a two-phase version of BoundlessBPE that produces an identical tokenization to the original BoundlessBPE training procedure presented by \citet{schmidt2025boundlessbytepairencoding}.
A high-level comparison of the two BoundlessBPE approaches and SuperBPE is given in \cref{fig:flowchart}.

\paragraph{The original procedure.}
The left of \cref{fig:flowchart} shows the original procedure.
The training data is pre-tokenized and used both as aggregated pretokens for ordinary BPE merges, and as raw documents of pretokens used to compute supermerges.
The single and pairwise merge counts are maintained separately for ordinary merges and supermerges.
A potential supermerge is \textit{unlocked}---becomes eligible for selection---only once both of its constituent pretokens have been created by ordinary merges, leading to a dynamic set of potential supermerges that grows as more ordinary merges are performed.
Training proceeds by choosing the available ordinary merge or supermerge with the highest count at each step, with counts and candidate supermerges updated accordingly.
The time to perform a merge is linear in the number of unique pretokens containing the merge pair, since they must all be considered.
Keeping raw document pretokens without aggregation causes extremely slow training, since each merge applies to more pretokens, and the long pretoken sequences must be scanned to locate all possible merges that apply.

\paragraph{The two-phase procedure.}
The right side of \cref{fig:flowchart} shows the two-phase procedure, which we describe box by box, relating each box to the corresponding step of the worked example in \cref{fig:example}. 
In Phase~1, the \emph{pre-tokenizer} splits the training data into pretokens and aggregates them by count (step 1 of \cref{fig:example}).
\emph{BPE training} trains an ordinary BPE model on these aggregated pretokens to the desired vocabulary size $|V|$ using $m = |V| - |\Sigma|$ merges,\footnote{$\Sigma$ is the initial vocabulary; it includes all single-byte tokens so all sequences can be tokenized without resorting to using \texttt{<unk>} for unseen tokens.} recording each merge together with the pair count at which BPE selected it.
The \emph{aggregation procedure} then takes this BPE model together with the pretokens and produces the supermerge candidates. Using the tokenization that the BPE model assigns to each pretoken, the procedure marks which pretokens reduce to a single token (step 2), forms maximal runs of adjacent eligible pretokens as candidates (step 3), and optionally splits long runs by $n$-gram frequency (step 4).
The resulting candidates are aggregated by frequency as the input to Phase~2.

In Phase~2 (\cref{alg:secondphase}) we replay the Phase~1 ordinary merges in the order BPE selected them---by non-increasing pair count---and interleave supermerges among them.
First, $\textsc{CountPairs}$ tallies all adjacent token pairs across the supermerge candidates.
At each step, $\textsc{BestPair}$ returns the highest-count adjacent pair among the candidates, and its count $c_s$ is compared against the count $c_r$ of the next ordinary merge in the replay.
If $c_s > c_r$, the supermerge is selected for addition: $\textsc{ApplyMerge}$ merges all occurrences of the pair in the candidates, adds the resulting token to the vocabulary, and updates the adjacent-pair counts, as the new token creates new adjacent pairs.
If $c_s \le c_r$, the ordinary merge's token is added to the vocabulary and the replay advances to the next ordinary merge.
SuperBPE (the right branch of \cref{fig:flowchart}) shares this same Phase~1 and differs only in Phase~2: rather than determining the split between ordinary merges and supermerges dynamically, it performs a fixed number $s$ of supermerges.
We return to this relationship in \cref{sec:relationship}.

\algrenewcommand{\algorithmicensure}{\textbf{Output:}}
\begin{algorithm}[ht]
\caption{BoundlessBPE Second-Phase Training}
\label{alg:secondphase}
\begin{algorithmic}[1]
\Require Initial single-byte vocabulary $\Sigma$
\Require First-phase BPE model $W$ with ordered merge list $M_W$ ($|M_W| + |\Sigma| = \textit{target\_size}$)
\Require Aggregated supermerge candidates with counts, from the aggregation procedure: $D = \{\textit{supermerge candidate} \mapsto \textit{count}\}$, where each supermerge candidate is a run of eligible, adjacent single-token pretokens
\Ensure Combined vocabulary $V$ containing both ordinary and superword tokens
\State $V \gets$ initial byte vocabulary $\Sigma$
\State $\textit{pair\_counts} \gets \Call{CountPairs}{D}$
\State $j \gets 0$ \Comment{Current index into $M_W$}
\While{$|V| < \textit{target\_size}$}
\State $(s_1, s_2), c_s \gets \Call{BestPair}{\textit{pair\_counts}}$ \Comment{Best supermerge candidate, or $c_s=0$ if none}
\State $(r_1, r_2), c_r \gets M_W[j]$ \Comment{Next ordinary merge from replay}
\If{$c_s > c_r$} \Comment{Supermerge wins (ties must go to ordinary)}
    \State $V \gets V \cup \{s_1 s_2\}$
    \State \Call{ApplyMerge}{$D, (s_1, s_2), \textit{pair\_counts}$} \Comment{Update counts and candidates}
\Else \Comment{Ordinary merge wins}
    \State $V \gets V \cup \{r_1 r_2\}$
    \State $j \gets j + 1$
\EndIf
\EndWhile
\end{algorithmic}
\end{algorithm}

\paragraph{Equivalence to the original procedure.}
Two-phase training produces the identical vocabulary and merges as the original single-pass procedure. The argument rests on three preconditions:
\begin{enumerate}
\item[(a)] Phase~1 trains the full $m = |V| - |\Sigma|$ ordinary merges, so the entire ordinary-merge sequence, in count order, is available for the replay in Phase~2.
\item[(b)] Ties between an ordinary merge and a supermerge are broken in favor of the ordinary merge (the strict inequality $c_s > c_r$ in \cref{alg:secondphase}).
\item[(c)] An ordinary merge combines two adjacent tokens only within a single pretoken; it never spans a pretoken boundary. A superword token spans such a boundary, so an ordinary merge never takes a superword token as one of its two inputs.
\end{enumerate}

Precondition~(c) is what decouples the two phases. A supermerge combines two whole pretokens that are each already a single token, so it does not change any ordinary-merge pairwise count; and by~(c) the resulting superword token is never subsequently taken as an input to an ordinary merge.
Consequently, the set of available ordinary merges and their counts are independent of any supermerge decision.
The ordinary merges chosen, and their order, are thus exactly those of a standalone BPE run to size $|V|$---which is what Phase~1 computes by precondition~(a).
This is why all ordinary merges can be decided in advance, without knowledge of any supermerge.

It remains to show that interleaving the replayed ordinary merges with supermerges by count (\cref{alg:secondphase}) makes the same choices as the original single-pass procedure, which at each step takes the available merge of either type with the highest count.
Assuming both procedures have applied the same merges so far, the next ordinary merge and its count agree by the argument above.
A supermerge is only ever selected once both of its parent pretokens are already single tokens.
Consider a supermerge with count $c_{ab}$ formed from two parent pretokens with counts $c_a$ and $c_b$; as with any BPE merge, $c_a \ge c_{ab}$ and $c_b \ge c_{ab}$.
While a parent still has a pending ordinary merge, that merge has count at least $c_{ab}$, so the next ordinary merge in the replay also has count at least $c_{ab}$ and is taken first---with ties going to the ordinary merge by precondition~(b).
This is exactly the condition under which the original procedure \textit{unlocks} the supermerge, so no explicit unlocking mechanism is needed.
Once both parents are single tokens, the supermerge's count is simply the number of adjacent occurrences of the two pretokens across the candidates, which is the same in both procedures.
Both therefore compare the same counts and make the same choice at every step.
Note that the converse does not hold: unlike SuperBPE, the ordinary-merge decisions do affect how many supermerges are ultimately chosen, since they determine which pretokens become single tokens and thus become eligible.

\subsection{Relationship Between BoundlessBPE and SuperBPE}
\label{sec:relationship}

This two-phase BoundlessBPE approach makes it easier to clearly see the relation between BoundlessBPE and SuperBPE, which has always had two phases.
In SuperBPE, the two phases are independent with a specified number of ordinary merges $r$ and supermerges $s$, where $|V| = r + s + |\Sigma|$.
In contrast, with BoundlessBPE, the full set of $m$ ordinary merges and merge counts from the first phase are used with \cref{alg:secondphase} in the second phase to simultaneously determine $r$ and $s$, eliminating the need to specify $s$ as a hyperparameter.
For example, suppose a BoundlessBPE model uses $r$ ordinary merges and $s$ supermerges.
A first-phase BPE model trained with $r$ merges, followed by a second-phase SuperBPE run with $s$ supermerges, will produce the exact same model as BoundlessBPE.
This theoretical observation has been verified experimentally in our implementation.\footnote{The vocabulary and merge rules are identical, but the integer token ID assignments will differ.}
We will use this \say{matching} SuperBPE model for computational results in the next section, to avoid tuning the number of supermerges $s$.
  
\section{Training Analysis}
\label{sec:computationalresults}

\begin{figure}[!ht]
    \centering
    \begin{minipage}[t]{0.54\linewidth}
        \centering
        \includegraphics[width=\linewidth]{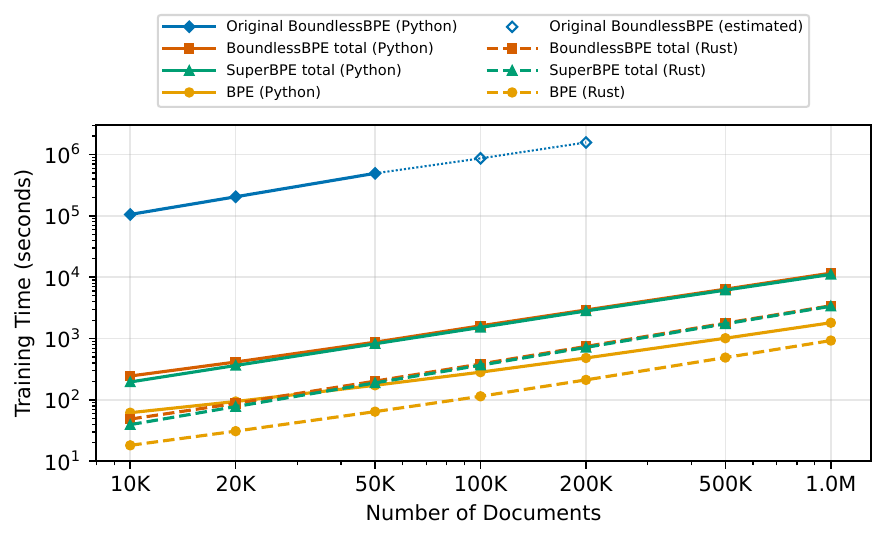}
        \caption{Average training time over 3 runs vs. number of documents, with a vocabulary size of 131,072. Error bars omitted as they were not visible on the log scale. The last two original BoundlessBPE points are estimated from smaller vocabulary sizes (see \cref{app:extrapolation}).}
        \label{fig:timing}
    \end{minipage}
    \hfill
    \begin{minipage}[t]{0.42\linewidth}
        \centering
        \includegraphics[width=\linewidth]{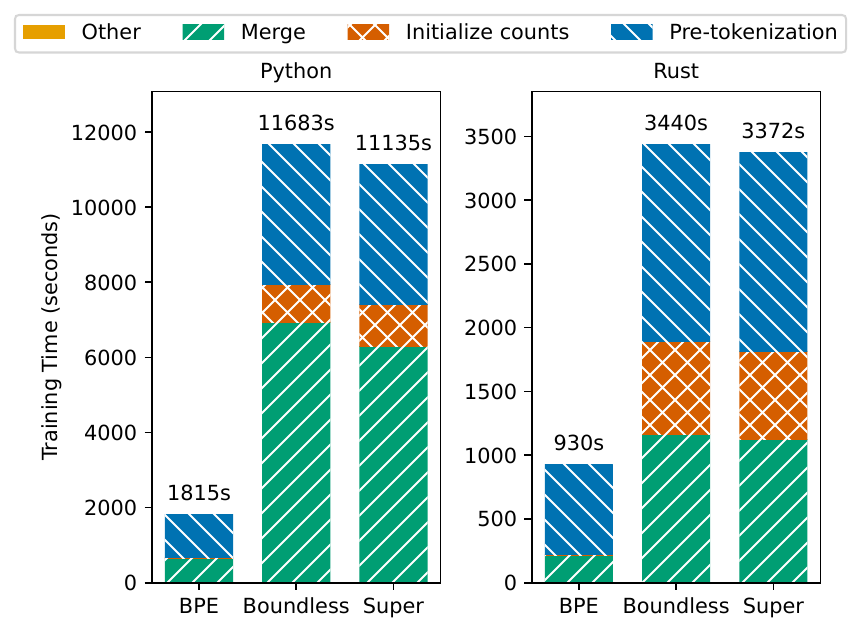}
         \caption{Average training time breakdown over 3 runs with 1M documents and vocabulary size 131,072. \emph{Initialize counts} includes computing initial single and pairwise counts and inserting them into a priority queue.}
        \label{fig:timingbreakdown}
    \end{minipage}
\end{figure}

\cref{fig:timing} shows the training time for BPE, BoundlessBPE, and SuperBPE over MiniPile documents, with a vocabulary size of 131,072.
Times for the original BoundlessBPE implementation are shown for comparison. 
These timings for BoundlessBPE and SuperBPE include the time of the first-phase BPE runs as well.
The timings for both Python and Rust implementations are shown. 

\cref{fig:timingbreakdown} gives the time breakdown for the 1M document runs in four categories.
For BPE, pre-tokenization consists only of regex splitting, which is already compiled C code in Python's \texttt{regex} library\footnote{\scriptsize{\url{https://github.com/mrabarnett/mrab-regex}}}.
The Rust speedup is therefore modest here.
In BoundlessBPE and SuperBPE, a majority of time is spent with the procedural code that performs the merges and updates the counts and candidates.
This procedural code benefits the most from moving to Rust, significantly reducing the time spent in merging relative to pre-tokenization.
With BoundlessBPE and SuperBPE, the fraction spent in initializing counts is also larger than for BPE, as there are up to $\sim$10$\times$ the number of unique supermerge candidates compared to the number of unique pretokens.
The time to run \cref{alg:secondphase} in BoundlessBPE is fairly small, accounting for the slight increase over SuperBPE times.
The overall speedup from Python to Rust is only about 3.4$\times$ at 1M documents, so the great majority of the more than 600-fold improvement comes from the aggregation procedure and two-phase decoupling, not the implementation language.

\begin{figure}[!ht]
    \centering
    \includegraphics[width=\linewidth]{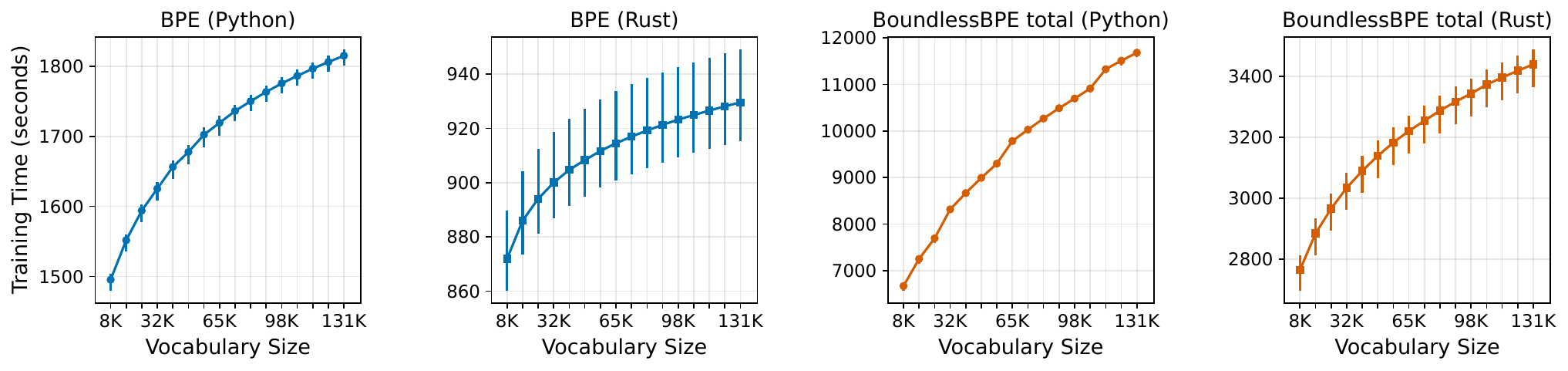}
    \caption{Average training time for BPE and BoundlessBPE (including both phases) over 3 runs vs. vocabulary size for 1M documents. Error bars give min/max values over the 3 runs. SuperBPE results are similar.}
    \label{fig:timingvsvocab}
\end{figure}

\cref{fig:timingvsvocab} shows the effect of vocabulary size on training time.
There is a significant setup time to pre-tokenize and compute the initial counts, and the early, very frequent merges are slow since they must be applied in many places, so the first 8,192 vocabulary entries account for much of the total time.
The curves are all sublinear, as the later merges apply to increasingly rarer pairs.
Once this initial cost is paid, expanding the vocabulary is relatively cheap.

\section{Inference Procedure}
\label{sec:inference}

Inference using a trained BoundlessBPE or SuperBPE model follows the data aggregation steps of \cref{fig:example}, with supermerges applied in step 5.
The Phase~1 ordinary BPE merges are first applied to each pretoken.
Then, the Phase~2 supermerges are applied to the resulting tokens of each supermerge candidate in turn.
The original BoundlessBPE implementation found the earliest merge rule across the entire document, then applied it to all pretokens.
This is correct, but very slow, since each pretoken must be checked for the rule, and many rules apply to only a small subset of pretokens.
It is much faster to find the earliest rule that applies to a given pretoken.
This is done by identifying all adjacent pairs that match a merge rule and applying the earliest (highest-count) rule.

\begin{figure}[!ht]
    \centering
    \includegraphics[width=\linewidth]{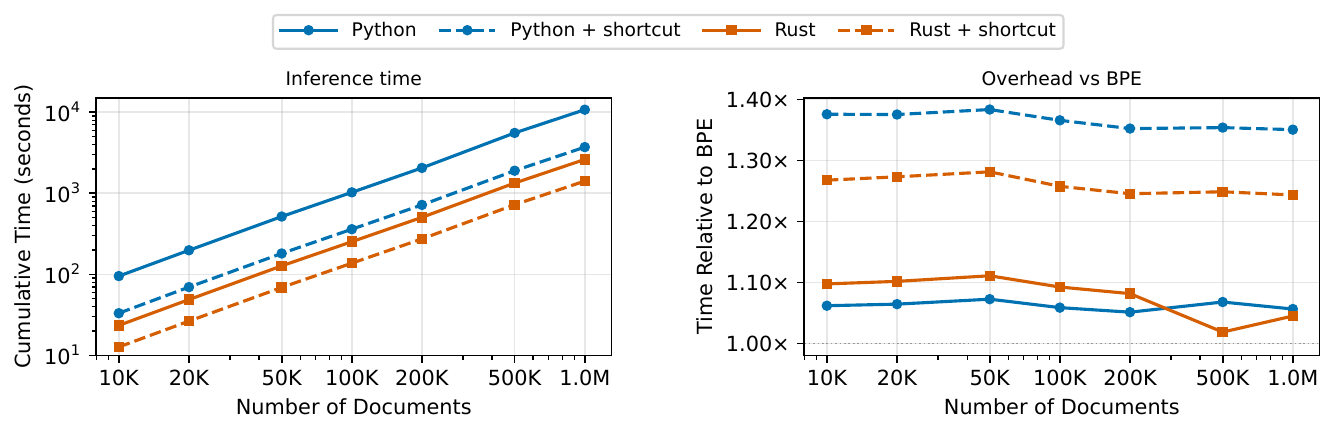}
    \caption{The left plot gives the inference time for BoundlessBPE vs. number of documents, both with and without the shortcut in Rust and Python. The right plot shows the time overhead of using BoundlessBPE vs. the first-phase word BPE. The shortcut provides a bigger boost to BPE than BoundlessBPE since the supermerge stage doesn't directly benefit. A ``matching'' SuperBPE model has the same merges as BoundlessBPE, so its inference times would be identical.}
    \label{fig:inferencetimes}
\end{figure}

With ordinary BPE, if a pretoken is contained in the vocabulary, it will end up tokenized as a single token.
Section 2 of \citet{schmidt2025boundlessbytepairencoding} shows this is the case for at least 90\% of pretokens.
This observation yields a simple but highly effective inference optimization applicable to any BPE implementation: rather than na{\"i}vely breaking all pretokens into bytes and applying the BPE merge rules, first check if the pretoken is in the vocabulary.
If it is, simply leave it as a single token and move on to the next pretoken.
This shortcut provides a significant 2.9$\times$ speedup for Python and a 1.85$\times$ speedup for Rust.

The cumulative time for inference over various numbers of MiniPile documents is given in \cref{fig:inferencetimes}. Using Rust compared to Python gives a 2.6$\times$ speedup when using the shortcut, and a 4.1$\times$ speedup without it.

\paragraph{Exporting to standard formats.}
A trained tokenizer can be exported to the Hugging Face and tiktoken byte-level BPE formats, so it can be used directly in LLM training or inference pipelines that expect one of these.
A first-phase word model exports exactly, producing a tokenization identical to ours.
A superword model has no exact representation, since neither format supports supermerges, but it can be approximated using the coarse-regex approach of the original SuperBPE implementation.
The approximation is close but not identical: fusing a run of pretokens under the coarse regex lets single-pass BPE apply some merges BoundlessBPE does not, differing on a meaningful fraction of documents.
\cref{app:export} gives the construction and characterizes the two ways the export can differ.

\section{Related Work}

Other approaches have been suggested for breaking up sequences of characters (or pretokens) beyond the greedy splitting approach of \cref{app:nonspacedelim}.  \citet{hu2025entropydrivenpretokenizationbytepairencoding} use an entropy-based approach to segment Chinese characters into smaller chunks, which could then be aggregated. The Byte Latent Transformer~\citep[BLT;][]{pagnoni-etal-2025-byte} also uses an entropy-based approach to find patches that is similar in spirit. \citet{goriely2025bytespaninformationdrivensubwordtokenisation} build on the BLT patching approach to design a stand-alone tokenizer.

Several works examine tokens that cross character boundaries. In these cases, partial bytes from adjacent characters are combined into the same token, which is thus not a valid UTF-8 character. \citet{firestone2025utf8plumbingbyteleveltokenizers} give the example of the Vedic Sanskrit word {\devanagarifont अग्निमीळे} (the opening word of the Rigveda) in the Devanagari script that is tokenized by \texttt{cl100k\_base}\footnote{\scriptsize {\url{https://github.com/openai/tiktoken}}} with two tokens crossing character boundaries. \citet{land2025bpestaysscriptstructured} report that the GPT-4o tokenizer has 874 tokens crossing character boundaries. They also trained a Thai language tokenizer with 42,831 tokens mixing full and partial characters, due to problematic initial merges. To alleviate this, they propose a constrained BPE merging strategy to prevent these merges from occurring. \citet{jang-etal-2025-improbable} focus on what they term \emph{incomplete tokens}---tokens with stray bytes attached that do not form complete characters---and study the problems caused by pairwise bigrams of such tokens.

\section{Conclusion}

We present a unified implementation of BoundlessBPE and SuperBPE that is significantly faster due to superword candidate aggregation and a two-phase decoupling of BoundlessBPE that is equivalent to the original. The BoundlessBPE implementation is more than 600 times faster (4.7 days to 603 seconds), and SuperBPE exhibits similar benefits. This enables training on larger datasets in reasonable time, which had been a barrier to adopting these promising approaches. The open-source implementation provides a straightforward reference Python implementation and a fast Rust implementation producing identical models. The current approach is single-threaded, and further efforts to support parallelization could provide additional speed improvements. We provide some preliminary results (\cref{app:nonspacedelim}) on using greedy left-to-right splitting to aggregate further in the context of supermerges. Further investigation is needed to determine how this greedy splitting affects downstream performance. We have not provided downstream analysis, but note that the extensive results by \citet{liu2025superbpespacetravellanguage} carry over, as these methods only speed up training without changing the resulting models.

\section*{Acknowledgments}
Thanks to Seth Ebner, Varshini Reddy, Adam Wiemerslage, and Michael Krumdick for their help with editing and improving the manuscript.
This research was supported in part by the Israel Science Foundation (grant No. 1166/23).


\bibliography{colm2026_conference,anthology-1}

\begin{thebibliography}{19}
\providecommand{\natexlab}[1]{#1}
\providecommand{\url}[1]{\texttt{#1}}
\expandafter\ifx\csname urlstyle\endcsname\relax
  \providecommand{\doi}[1]{doi: #1}\else
  \providecommand{\doi}{doi: \begingroup \urlstyle{rm}\Url}\fi

\bibitem[Agrawal \& Srikant(1994)Agrawal and Srikant]{10.5555/645920.672836}
Rakesh Agrawal and Ramakrishnan Srikant.
\newblock Fast algorithms for mining association rules in large databases.
\newblock In \emph{Proceedings of the 20th International Conference on Very Large Data Bases}, VLDB '94, pp.\  487–499, San Francisco, CA, USA, 1994. Morgan Kaufmann Publishers Inc.
\newblock ISBN 1558601538.

\bibitem[Anthropic(2026)]{anthropic2026claude}
Anthropic.
\newblock Claude {O}pus 4.6, 2026.
\newblock URL \url{https://www.anthropic.com/claude}.

\bibitem[Berberich \& Bedathur(2013)Berberich and Bedathur]{10.1145/2452376.2452389}
Klaus Berberich and Srikanta Bedathur.
\newblock Computing n-gram statistics in mapreduce.
\newblock In \emph{Proceedings of the 16th International Conference on Extending Database Technology}, EDBT '13, pp.\  101–112, New York, NY, USA, 2013. Association for Computing Machinery.
\newblock ISBN 9781450315975.
\newblock \doi{10.1145/2452376.2452389}.
\newblock URL \url{https://doi.org/10.1145/2452376.2452389}.

\bibitem[Chizhov et~al.(2024)Chizhov, Arnett, Korotkova, and Yamshchikov]{chizhov-etal-2024-bpe}
Pavel Chizhov, Catherine Arnett, Elizaveta Korotkova, and Ivan~P. Yamshchikov.
\newblock {BPE} gets picky: Efficient vocabulary refinement during tokenizer training.
\newblock In Yaser Al-Onaizan, Mohit Bansal, and Yun-Nung Chen (eds.), \emph{Proceedings of the 2024 Conference on Empirical Methods in Natural Language Processing}, pp.\  16587--16604, Miami, Florida, USA, November 2024. Association for Computational Linguistics.
\newblock \doi{10.18653/v1/2024.emnlp-main.925}.
\newblock URL \url{https://aclanthology.org/2024.emnlp-main.925/}.

\bibitem[Firestone et~al.(2025)Firestone, Ugare, Singh, and Misailovic]{firestone2025utf8plumbingbyteleveltokenizers}
Preston Firestone, Shubham Ugare, Gagandeep Singh, and Sasa Misailovic.
\newblock {UTF-8} plumbing: Byte-level tokenizers unavoidably enable {LLMs} to generate ill-formed {UTF-8}, 2025.
\newblock URL \url{https://arxiv.org/abs/2511.05578}.

\bibitem[Gage(1994)]{gage1994new}
Philip Gage.
\newblock A new algorithm for data compression.
\newblock \emph{The C Users Journal}, 12\penalty0 (2):\penalty0 23--38, 1994.

\bibitem[Goriely et~al.(2025)Goriely, Salhan, Lesci, Cheng, and Buttery]{goriely2025bytespaninformationdrivensubwordtokenisation}
Zébulon Goriely, Suchir Salhan, Pietro Lesci, Julius Cheng, and Paula Buttery.
\newblock {ByteSpan}: Information-driven subword tokenisation, 2025.
\newblock URL \url{https://arxiv.org/abs/2506.18639}.

\bibitem[Hu et~al.(2025)Hu, Liang, Zhao, Geuter, Reddy, Schmidt, and Tanner]{hu2025entropydrivenpretokenizationbytepairencoding}
Yifan Hu, Frank Liang, Dachuan Zhao, Jonathan Geuter, Varshini Reddy, Craig~W. Schmidt, and Chris Tanner.
\newblock Entropy-driven pre-tokenization for byte-pair encoding, 2025.
\newblock URL \url{https://arxiv.org/abs/2506.15889}.

\bibitem[Jang et~al.(2025)Jang, Lee, Chung, Park, and Shin]{jang-etal-2025-improbable}
Eugene Jang, Kimin Lee, Jin-Woo Chung, Keuntae Park, and Seungwon Shin.
\newblock Improbable bigrams expose vulnerabilities of incomplete tokens in byte-level tokenizers.
\newblock In Christos Christodoulopoulos, Tanmoy Chakraborty, Carolyn Rose, and Violet Peng (eds.), \emph{Proceedings of the 2025 Conference on Empirical Methods in Natural Language Processing}, pp.\  18209--18216, Suzhou, China, November 2025. Association for Computational Linguistics.
\newblock ISBN 979-8-89176-332-6.
\newblock \doi{10.18653/v1/2025.emnlp-main.919}.
\newblock URL \url{https://aclanthology.org/2025.emnlp-main.919/}.

\bibitem[Kaddour(2023)]{kaddour2023minipilechallengedataefficientlanguage}
Jean Kaddour.
\newblock The {MiniPile} challenge for data-efficient language models, 2023.
\newblock URL \url{https://arxiv.org/abs/2304.08442}.

\bibitem[Land \& Arnett(2025)Land and Arnett]{land2025bpestaysscriptstructured}
Sander Land and Catherine Arnett.
\newblock {BPE} stays on {SCRIPT}: Structured encoding for robust multilingual pretokenization, 2025.
\newblock URL \url{https://arxiv.org/abs/2505.24689}.

\bibitem[Liu et~al.(2025)Liu, Hayase, Hofmann, Oh, Smith, and Choi]{liu2025superbpespacetravellanguage}
Alisa Liu, Jonathan Hayase, Valentin Hofmann, Sewoong Oh, Noah~A. Smith, and Yejin Choi.
\newblock {SuperBPE}: Space travel for language models, 2025.
\newblock URL \url{https://arxiv.org/abs/2503.13423}.

\bibitem[Nguyen et~al.(2024)Nguyen, Nguyen, Lai, Man, Ngo, Dernoncourt, Rossi, and Nguyen]{nguyen-etal-2024-culturax}
Thuat Nguyen, Chien~Van Nguyen, Viet~Dac Lai, Hieu Man, Nghia~Trung Ngo, Franck Dernoncourt, Ryan~A. Rossi, and Thien~Huu Nguyen.
\newblock {C}ultura{X}: A cleaned, enormous, and multilingual dataset for large language models in 167 languages.
\newblock In Nicoletta Calzolari, Min-Yen Kan, Veronique Hoste, Alessandro Lenci, Sakriani Sakti, and Nianwen Xue (eds.), \emph{Proceedings of the 2024 Joint International Conference on Computational Linguistics, Language Resources and Evaluation (LREC-COLING 2024)}, pp.\  4226--4237, Torino, Italia, May 2024. ELRA and ICCL.
\newblock URL \url{https://aclanthology.org/2024.lrec-main.377/}.

\bibitem[Pagnoni et~al.(2025)Pagnoni, Pasunuru, Rodriguez, Nguyen, Muller, Li, Zhou, Yu, Weston, Zettlemoyer, Ghosh, Lewis, Holtzman, and Iyer]{pagnoni-etal-2025-byte}
Artidoro Pagnoni, Ramakanth Pasunuru, Pedro Rodriguez, John Nguyen, Benjamin Muller, Margaret Li, Chunting Zhou, Lili Yu, Jason~E Weston, Luke Zettlemoyer, Gargi Ghosh, Mike Lewis, Ari Holtzman, and Srini Iyer.
\newblock Byte latent transformer: Patches scale better than tokens.
\newblock In Wanxiang Che, Joyce Nabende, Ekaterina Shutova, and Mohammad~Taher Pilehvar (eds.), \emph{Proceedings of the 63rd Annual Meeting of the Association for Computational Linguistics (Volume 1: Long Papers)}, pp.\  9238--9258, Vienna, Austria, July 2025. Association for Computational Linguistics.
\newblock ISBN 979-8-89176-251-0.
\newblock \doi{10.18653/v1/2025.acl-long.453}.
\newblock URL \url{https://aclanthology.org/2025.acl-long.453/}.

\bibitem[Reddy et~al.(2025)Reddy, Schmidt, Pinter, and Tanner]{reddy2025enoughdiminishingreturnstokenization}
Varshini Reddy, Craig~W. Schmidt, Yuval Pinter, and Chris Tanner.
\newblock How much is enough? the diminishing returns of tokenization training data, 2025.
\newblock URL \url{https://arxiv.org/abs/2502.20273}.

\bibitem[Schmidt et~al.(2025)Schmidt, Reddy, Tanner, and Pinter]{schmidt2025boundlessbytepairencoding}
Craig~W. Schmidt, Varshini Reddy, Chris Tanner, and Yuval Pinter.
\newblock Boundless byte pair encoding: Breaking the pre-tokenization barrier, 2025.
\newblock URL \url{https://arxiv.org/abs/2504.00178}.

\bibitem[Sennrich et~al.(2016)Sennrich, Haddow, and Birch]{sennrich-etal-2016-improving}
Rico Sennrich, Barry Haddow, and Alexandra Birch.
\newblock Improving neural machine translation models with monolingual data.
\newblock In Katrin Erk and Noah~A. Smith (eds.), \emph{Proceedings of the 54th Annual Meeting of the Association for Computational Linguistics (Volume 1: Long Papers)}, pp.\  86--96, Berlin, Germany, August 2016. Association for Computational Linguistics.
\newblock \doi{10.18653/v1/P16-1009}.
\newblock URL \url{https://aclanthology.org/P16-1009/}.

\bibitem[Singh \& Strouse(2024)Singh and Strouse]{singh2024tokenizationcountsimpacttokenization}
Aaditya~K. Singh and DJ~Strouse.
\newblock Tokenization counts: the impact of tokenization on arithmetic in frontier llms, 2024.
\newblock URL \url{https://arxiv.org/abs/2402.14903}.

\bibitem[Zouhar et~al.(2023)Zouhar, Meister, Gastaldi, Du, Sachan, and Cotterell]{zouhar-etal-2023-tokenization}
Vil{\'e}m Zouhar, Clara Meister, Juan Gastaldi, Li~Du, Mrinmaya Sachan, and Ryan Cotterell.
\newblock Tokenization and the noiseless channel.
\newblock In Anna Rogers, Jordan Boyd-Graber, and Naoaki Okazaki (eds.), \emph{Proceedings of the 61st Annual Meeting of the Association for Computational Linguistics (Volume 1: Long Papers)}, pp.\  5184--5207, Toronto, Canada, July 2023. Association for Computational Linguistics.
\newblock \doi{10.18653/v1/2023.acl-long.284}.
\newblock URL \url{https://aclanthology.org/2023.acl-long.284/}.

\end{thebibliography}
\bibliographystyle{colm2026_conference}

\appendix

\section{LLM Use Disclosure}

We used Claude Code Opus 4.6 \citep{anthropic2026claude} to automatically translate the Python reference implementation into the faster Rust version. Correctness of the training code was verified by comparing that the Python and Rust codes produced completely identical models over a range of document sizes.  Inference was verified correct by ensuring the tokenization of all training documents was identical between both implementations. The LLM was also used to write the Python scripts to generate the figures in this paper, and to improve paper wording at a paragraph level.

\crefalias{section}{appendix}
\section{Pre-tokenization Regular Expressions}
\label{app:pretokenization}
Pre-tokenization is the process of applying a regular expression (regex) to split a document into smaller parts called pretokens, which are then each tokenized independently.
We use the modern \texttt{GPT4O\_REGEX} of \cref{lst:regex} for pre-tokenization, originally introduced for GPT-4o.\footnote{From {\scriptsize\url{https://github.com/openai/tiktoken/blob/4560a889/tiktoken_ext/openai_public.py\#L101-L114}}.}
The seven branches in this regex are joined with the alternation operator \texttt{|}, and the regex engine attempts to match each branch in turn at the current position, accepting the first branch that matches. The order of the alternatives is therefore significant, since an earlier branch takes precedence over any later branch that could also have matched. \cref{tab:regex-branches} gives examples of each branch.

\begin{figure}[ht!]
\begin{lstlisting}
import regex as re

GPT4O_REGEX = "|".join([
    r"[^\r\n\p{L}\p{N}]?[\p{Lu}\p{Lt}\p{Lm}\p{Lo}\p{M}]*\
[\p{Ll}\p{Lm}\p{Lo}\p{M}]+(?i:'s|'t|'re|'ve|'m|'ll|'d)?",   # 1
    r"[^\r\n\p{L}\p{N}]?[\p{Lu}\p{Lt}\p{Lm}\p{Lo}\p{M}]+\
[\p{Ll}\p{Lm}\p{Lo}\p{M}]*(?i:'s|'t|'re|'ve|'m|'ll|'d)?",   # 2
    r"\p{N}{1,3}",                                          # 3
    r" ?[^\s\p{L}\p{N}]+[\r\n/]*",                          # 4
    r"\s*[\r\n]+",                                          # 5
    r"\s+(?!\S)",                                           # 6
    r"\s+",                                                 # 7
])
\end{lstlisting}
\captionof{listing}{The GPT-4o pre-tokenization regex, with its seven branches numbered.}
\label{lst:regex}
\end{figure}


\begin{table}[ht!]
\centering
\small
\begin{tabular}{@{}c p{2.4in} l l@{}}
\toprule
\# & Matches & Input & Output \\
\midrule
\multirow{6}{*}{1} & \multirow{6}{*}{\parbox[t]{2.4in}{Optional space or punctuation, optional upper or uncased letters followed by one or more lowercase letters. Breaks before an uppercase letter that follows a lowercase one. Keeps trailing modifiers (accents).}} & \texttt{\sq ~Hello~world\sq} & \texttt{[[\sq ~Hello\sq],[\sq ~world\sq]]} \\
  &  & \texttt{\sq ~camelCase\sq}  & \texttt{[[\sq ~camel\sq],[\sq Case\sq]]} \\
  &  & \texttt{\sq FOObar\sq}     & \texttt{[[\sq FOObar\sq]]} \\
  &  & \texttt{\sq (can\sq t\sq}  & \texttt{[[\sq (can\sq t\sq]]} \\
  &  & \texttt{\sq caf\'e\sq}     & \texttt{[[\sq caf\'e\sq ]]} \\
  &  & \texttt{\sq }{\cjkfont 你好}\texttt{\sq}            & \texttt{[[\sq }{\cjkfont 你好}\texttt{\sq ]]} \\
\midrule
2 & Letters excluding lowercase (acronym).            & \texttt{\sq HTML~CSS\sq} & \texttt{[[\sq HTML\sq ],[\sq ~CSS\sq]]} \\
\midrule
3 & Digit runs of at most three digits.                  & \texttt{\sq 1234567 \sq }        & \texttt{[[\sq 123\sq ],[\sq 456\sq ],[\sq 7\sq ]]} \\
\midrule
4 & Punctuation or symbol run, keeping any trailing line endings. & \texttt{\sq .~!\$\%\$~\%\textbackslash n\sq} & \texttt{[[\sq .\sq ],[\sq ~!\$\%\$\sq],[\sq ~\%\textbackslash n\sq]]} \\
\midrule
5 & Whitespace ending in one or more line endings; spans intervening line endings. & \texttt{\sq ~~\textbackslash n~~~~\textbackslash n\textbackslash n\sq} & \texttt{[[\sq ~~\textbackslash n~~~~\textbackslash n\textbackslash n\sq]]} \\
\midrule
6 & Whitespace run, reserving its final space for the next word. (The \texttt{\sq a\sq } and \texttt{\sq ~b\sq } match branch 1.) & \texttt{\sq a~~~b\sq} & \texttt{[[\sq a\sq ],[\sq ~~\sq],[\sq ~b\sq]]} \\
\midrule
7 & Any remaining whitespace, such as a space before a number (matching branch 3). & \texttt{\sq ~12\sq} & \texttt{[[\sq ~\sq],[\sq 12\sq ]]} \\
\bottomrule
\end{tabular}
\caption{Behavior of each branch of \texttt{GPT4O\_REGEX} (\cref{lst:regex}). The seven alternatives are tried in order; the first to match at the current position wins.}
\label{tab:regex-branches}
\end{table}

Branches~1 and~2 both match words. Branch~1 matches an optional run of uppercase or uncased letters followed by one or more lowercase letters. Once the required lowercase run has begun, the match ends at the next uppercase letter. A word is therefore divided only before an uppercase letter that immediately follows a lowercase letter, as between \texttt{\sq camel\sq } and \texttt{\sq Case\sq }. Because branch~1 requires a lowercase letter, a word written entirely in uppercase cannot match it. Branch~2 handles this remaining case, matching a run of uppercase letters so that an acronym such as \texttt{\sq HTML\sq } or \texttt{\sq ~CSS\sq } is kept whole. Each branch begins with an optional leading character that is neither a letter, digit, nor line ending, so that a preceding space, or the parenthesis in \texttt{\sq (can\sq t\sq}, is attached to the word that follows it. Each branch also ends with an optional suffix from a fixed list of English contractions, which keeps \texttt{can\sq t} intact, and permits trailing combining marks (\texttt{\textbackslash p\{M\}}), so that \texttt{caf\'e} remains a single chunk whether its accent is a pre-composed code point or a separate combining mark. Because both branches also match uncased letters (\texttt{\textbackslash p\{Lo\}}), a run of Han characters such as {\cjkfont 你好} is matched by branch~1. This behavior motivates the script-specific regex of \cref{lst:scriptspecificregex}, which instead assigns each such character to its own pretoken.

Branches 3 and 4 deal with numbers and punctuation. Branch~3 groups digits three at a time left-to-right, so that \texttt{\sq 1234567\sq } is divided into \texttt{[[\sq 123\sq ],[\sq 456\sq ],[\sq 7\sq ]]}.\footnote{ \citet{singh2024tokenizationcountsimpacttokenization} argue that LLMs perform better with digits in groups of three right-to-left, such as \texttt{[[\sq 1\sq ],[\sq 234\sq ],[\sq 567\sq ]]}. See Appendix A of \citet{schmidt2025boundlessbytepairencoding} for one way to implement this with a regex.} Branch~4 matches a run of punctuation or symbols, with an optional leading space, and retains any immediately following line ending, so that \texttt{\%\textbackslash n} is kept together. 

The final three branches divide whitespace. Branch~5 matches a run that ends in a line ending, spanning any intervening spaces and line endings, as in \texttt{\sq ~~\textbackslash n~~~~\textbackslash n\textbackslash n\sq}. Branch~6 applies the negative lookahead \texttt{(?!\textbackslash S)} to stop one space short of a following word, reserving that space for the word, so that \texttt{\sq a~~~b\sq } is divided into \texttt{[[\sq a\sq ],[\sq ~~\sq],[\sq ~b\sq]]}. This lookahead plays a crucial role in allowing leading spaces in words. Branch~7 matches any remaining whitespace not consumed by the preceding branches, such as the space before a number in \texttt{[[\sq ~\sq],[\sq 12\sq ]]}, which arises because branch~3 has no leading-space prefix of its own.

Appendix A of \citet{schmidt2025boundlessbytepairencoding} has extensive further discussion of other commonly used regex. 

We implement an optional, script-specific regular expression that is useful for non-space-delimited scripts.
Consider the impact of text in languages such as Chinese or Japanese on supermerges.\footnote{We use Han, Hiragana, Katakana, Thai, Myanmar, Khmer, and Lao as our non-space-delimited scripts. Chinese uses Han exclusively while Japanese uses Han, Hiragana, and Katakana. Myanmar is the script used for the Burmese language.}
Supermerges usually combine space-delimited words, but since these scripts have no spaces, the entire run of letters will be a single pretoken, leaving no consecutive pretokens eligible for supermerges.
A second, independent issue is that BPE can form tokens composed of partial bytes from two or more characters, producing invalid UTF-8 sequences~\citep{firestone2025utf8plumbingbyteleveltokenizers,land2025bpestaysscriptstructured,jang-etal-2025-improbable}.
This is especially likely with the 3-byte UTF-8 characters used by these scripts.
To address both problems, we split each character into its own 3-byte pretoken using a script-specific regex.
This both prevents partial-byte tokens and creates a sequence of single-character pretokens eligible for supermerges, allowing characters to combine as their counts permit.

In order to know when to apply this regular expression, we first split a document into chunks that contain only a single script.
This splitting also prevents pretokens from containing multiple scripts, such as \texttt{\sq {\cjkfont 德国}HYDAC{\cjkfont 电磁球阀}\sq } containing both Han and Latin.\footnote{This example is from the CulturaX multilingual dataset \citep{nguyen-etal-2024-culturax}. It anecdotally appears that mixing of English and Chinese is fairly common in the Chinese portion of the dataset.}
The splitting process can be done in linear time, since each Unicode code point is statically assigned to the category \textsc{Common}, \textsc{Inherited}, or one of 172 scripts such as \textsc{Latin} or \textsc{Han}.\footnote{The script assignments are from {\scriptsize\url{https://www.unicode.org/Public/17.0.0/ucd/Scripts.txt}}. We treat Unknown code points as \textsc{Common}.}
The example document from \cref{fig:example} contains a mix of \textsc{Latin}, \textsc{Han}, and \textsc{Common} code points such as space and punctuation:

\begin{center} \texttt{\sq Tokenization of the multilingual {\cjkfont 德国}HYDAC{\cjkfont 电磁球阀} can be hard.\sq }.
\end{center}

With script-specific splitting this would become: 

\begin{center} \texttt{[\sq Tokenization of the multilingual\sq,\sq ~{\cjkfont 德国}\sq ,\sq HYDAC\sq ,\sq {\cjkfont 电磁球阀},\sq can be hard.\sq ]}.
\end{center}

We split such that \textsc{Common} code points are attached to the script-specific chunk that follows, so, in particular, this results in leading spaces like \texttt{\sq ~}{\cjkfont 德国}\texttt{\sq } and \texttt{\sq ~can be hard.\sq }.
\textsc{Inherited} code points are combining modifiers such as diacritical marks, so they are kept at the end of the script-specific character they modify.

\begin{figure}[ht!]
\begin{lstlisting}
SCRIPT_SPECIFIC_GPT4O_REGEX = "|".join([
    r" ?\p{L}\p{M}*",
    r"\p{N}{1,3}",
    r" ?[^\s\p{L}\p{N}]+[\r\n/]*",
    r"\s*[\r\n]+",
    r"\s+(?!\S)",
    r"\s+",
])

# Default scripts that benefit from character-by-character tokenization
DEFAULT_SCRIPT_SPECIFIC_SCRIPTS = [
    'Han', 'Hiragana', 'Katakana', 'Thai',
    'Myanmar', 'Khmer', 'Lao'
]

\end{lstlisting}
\captionof{listing}{The script-specific variant for non-space-delimited languages.  By default it applies to the Han, Hiragana, Katakana, Thai, Myanmar, Khmer, and Lao scripts.  It can be applied to other scripts as desired.}
\label{lst:scriptspecificregex}
\end{figure}

For non-space-delimited scripts we use the \texttt{SCRIPT\_SPECIFIC\_GPT4O\_REGEX} in \cref{lst:scriptspecificregex}, a modified version that replaces the first two branches of the regex (which match words) with \texttt{"~?\textbackslash p\{L\}\textbackslash p\{M\}*"}.
This pattern matches single Unicode letters, with an optional leading space and trailing combining modifiers.
We apply the script-specific form to the non-space-delimited scripts in \texttt{DEFAULT\_SCRIPT\_SPECIFIC\_SCRIPTS}. 
The benefit of splitting single multi-byte characters into pretokens is that it prevents BPE from forming tokens that span character boundaries---tokens that combine a trailing fragment of one character with a leading fragment of the next, which are not valid UTF-8.
Later, supermerges can recombine the full characters into multi-character phrases.

In step 1 of \cref{fig:example}, each single-script chunk is pre-tokenized separately using a regex.
For space-delimited scripts we use the GPT-4o regular expression of \cref{lst:regex}, and for non-space-delimited scripts we use the script-specific regex of \cref{lst:scriptspecificregex}.

The computational results in this paper use the script-specific pre-tokenization of \cref{lst:scriptspecificregex} for the non-space-delimited scripts, together with the standard \texttt{GPT4O\_REGEX} of \cref{lst:regex} for all other text.
Because the MiniPile training corpus is almost entirely in English, script-specific splitting affects very little of the data.
Nevertheless, the separate behavior was enabled throughout.

\section{Exporting Models to Hugging Face and tiktoken}
\label{app:export}
Beyond our native model format, our implementation can export a trained model to either the Hugging Face\footnote{\scriptsize\url{https://github.com/huggingface/tokenizers}} or tiktoken\footnote{\scriptsize\url{https://github.com/openai/tiktoken}} format, so a model trained with our code can be directly deployed in an LLM training or inference pipeline using these formats.
Both formats are regular single-pass byte-level BPE, so we restrict export to models trained with no PickyBPE deletions and no script-specific pre-tokenization (\cref{app:pretokenization}), as neither is supported in these formats.
A first-phase BPE word model maps directly onto both formats, and the exported tokenizer produces identical tokenizations to ours on all inputs.
The remainder of this appendix discusses how to approximately export BoundlessBPE and SuperBPE models.

A superword model has no exact representation in these regular BPE formats, but it can be approximated.
Using the same approach as the original SuperBPE implementation,\footnote{\scriptsize\url{https://superbpe.github.io/}} we emit the model as a single-pass byte-level BPE with ordinary merges ordered before supermerges (so all within-word merges complete before any cross-word merge), and pair it with a modified pre-tokenization regex.
While the normal \emph{fine} regex used to train the superword model splits text into individual word pretokens, we switch to a related \emph{coarse} regex that keeps a maximal run of consecutive, supermerge-eligible word pretokens together as one pretoken.
Plain single-pass BPE over that fused pretoken first performs the ordinary within-word merges and then, because the whole run is one pretoken, the cross-word supermerges, which approximately reproduces the two phases of BoundlessBPE or SuperBPE without any special machinery.
This is the ungated behavior of the original SuperBPE rather than the whole-pretoken variant we use in this paper, which is exactly why the export only approximates BoundlessBPE: because whitespace no longer blocks merges, single-pass BPE can combine partial words across the original pretoken boundaries, something BoundlessBPE never does.
The two differences characterized below are precisely these cross-boundary merges.

The coarse regex is obtained from the fine one by taking its word-matching branches and combining them into a single repeatable alternation using the \texttt{|} operator.
If those branches from the fine regex are \texttt{a} and \texttt{b}, then in the coarse regex they appear as one repeated group \texttt{(?:a|b)+}, so that a run of adjacent words matches as a single coarse pretoken.
The remaining number, punctuation, and whitespace branches are left unchanged.
This construction works for any fine regex, including the default \texttt{GPT4O\_REGEX} of \cref{lst:regex}.
We prefer to avoid punctuation in superwords, so we use the \texttt{GPT4O\_EXPORT\_REGEX} and \texttt{GPT4O\_COARSE\_REGEX} of \cref{lst:exportregex}.
Their word branches begin with an optional leading space rather than the optional leading punctuation character of \texttt{GPT4O\_REGEX}, so a preceding punctuation mark is not pulled into the word.\footnote{The word branches also change how apostrophes are handled, replacing GPT-4o's fixed list of English contraction suffixes with an apostrophe that must sit between letters, which is more general and less English-centric.}
A model is trained with the fine regex and exported with the coarse one.

\begin{figure}[ht!]
\begin{lstlisting}
# Fine regex used to TRAIN an exportable superword model.
# Branches 3-7 are identical to GPT4O_REGEX of Listing 1.
GPT4O_EXPORT_REGEX = "|".join([
    r" ?[\p{Lu}\p{Lt}\p{Lm}\p{Lo}\p{M}]*\
[\p{Ll}\p{Lm}\p{Lo}\p{M}]+(?:['@\textquoteright@](?:\p{L}\p{M}*)+)*",       # 1
    r" ?[\p{Lu}\p{Lt}\p{Lm}\p{Lo}\p{M}]+\
[\p{Ll}\p{Lm}\p{Lo}\p{M}]*(?:['@\textquoteright@](?:\p{L}\p{M}*)+)*",       # 2
    r"\p{N}{1,3}",                                          # 3
    r" ?[^\s\p{L}\p{N}]+[\r\n/]*",                          # 4
    r"\s*[\r\n]+",                                          # 5
    r"\s+(?!\S)",                                           # 6
    r"\s+",                                                 # 7
])

# Coarse regex used to EXPORT: word branches 1 and 2 are combined
# into one repeated group, so a run of words becomes ONE pretoken.
# Branches 3-7 are unchanged.
GPT4O_COARSE_REGEX = "|".join([
    r"(?: ?[\p{Lu}\p{Lt}\p{Lm}\p{Lo}\p{M}]*\
[\p{Ll}\p{Lm}\p{Lo}\p{M}]+(?:['@\textquoteright@](?:\p{L}\p{M}*)+)*\
| ?[\p{Lu}\p{Lt}\p{Lm}\p{Lo}\p{M}]+\
[\p{Ll}\p{Lm}\p{Lo}\p{M}]*(?:['@\textquoteright@](?:\p{L}\p{M}*)+)*)+",     # 1+2
    r"\p{N}{1,3}",                                          # 3
    r" ?[^\s\p{L}\p{N}]+[\r\n/]*",                          # 4
    r"\s*[\r\n]+",                                          # 5
    r"\s+(?!\S)",                                           # 6
    r"\s+",                                                 # 7
])
\end{lstlisting}
\captionof{listing}{The fine/coarse regex pair for exporting superword models. Compared with \texttt{GPT4O\_REGEX} (\cref{lst:regex}), the word branches allow only a leading space and change how contractions are handled; the coarse regex additionally groups the two word branches so a run of words becomes one pretoken.}
\label{lst:exportregex}
\end{figure}

We emphasize that this approach does not give identical tokenizations to BoundlessBPE.
In BoundlessBPE two constraints hold at inference: a supermerge combines only whole, single-token pretokens; and no ordinary merge ever crosses a pretoken boundary.
When the coarse regex fuses a run of word pretokens into one pretoken, the original pretoken boundaries are lost, and either constraint can then be violated.
The export approach will thus sometimes perform a supermerge using only part of an original pretoken.
Consider the example \texttt{\sq ~are notes\sq}. The fine regex splits it into two pretokens \texttt{[\sq ~are\sq],[\sq ~notes\sq]}, and after within-word merges these become \texttt{[\sq ~are\sq]} and \texttt{[\sq ~not\sq,\sq es\sq]}. Suppose the supermerge of \texttt{\sq ~are\sq} and \texttt{\sq ~not\sq} is in the vocabulary. BoundlessBPE will leave this as three tokens, because \texttt{\sq ~not\sq} is not a whole single-token pretoken and is therefore ineligible for a supermerge. The export approach sees one fused coarse pretoken \texttt{\sq ~are~notes\sq}, which after ordinary merges would be \texttt{[\sq ~are\sq,\sq ~not\sq,\sq es\sq]}. It then applies the supermerge, giving \texttt{[\sq ~are~not\sq,\sq es\sq]}, absorbing part of the second word.
This is the more common of the two differences: roughly a quarter of documents contained at least one such supermerge.

The export approach will also sometimes perform an ordinary merge across a pretoken boundary:
within a fused coarse pretoken, the bytes on either side of the original pretoken boundary become adjacent, so an ordinary merge can combine them.
This can even split a multi-byte UTF-8 character down the middle: for example, in the Cyrillic-script identifier \texttt{\cyr{ФичаДляПроверкиМетода}}, the export approach merges the \texttt{\cyr{а}} ending \texttt{\cyr{Фича}} with the first byte of the \texttt{\cyr{Д}} beginning \texttt{\cyr{Для}}, leaving its second byte in the next token.
Such tokens combine fragments of adjacent characters and are not valid UTF-8.
This case is fairly rare, with 24 occurrences over the 1M MiniPile documents, always in camelCase identifiers in Cyrillic.
However, it would likely appear more frequently in multilingual datasets.

For applications that need exact BoundlessBPE behavior we recommend using our tokenizer directly; the export approach allows deployment in standard pipelines where these two tokenization differences are acceptable.

\section{Left-to-right Greedy Splitting of Pretoken n-grams}
\label{app:nonspacedelim}

\begin{figure}[!ht]
    \centering
    \includegraphics[width=\linewidth]{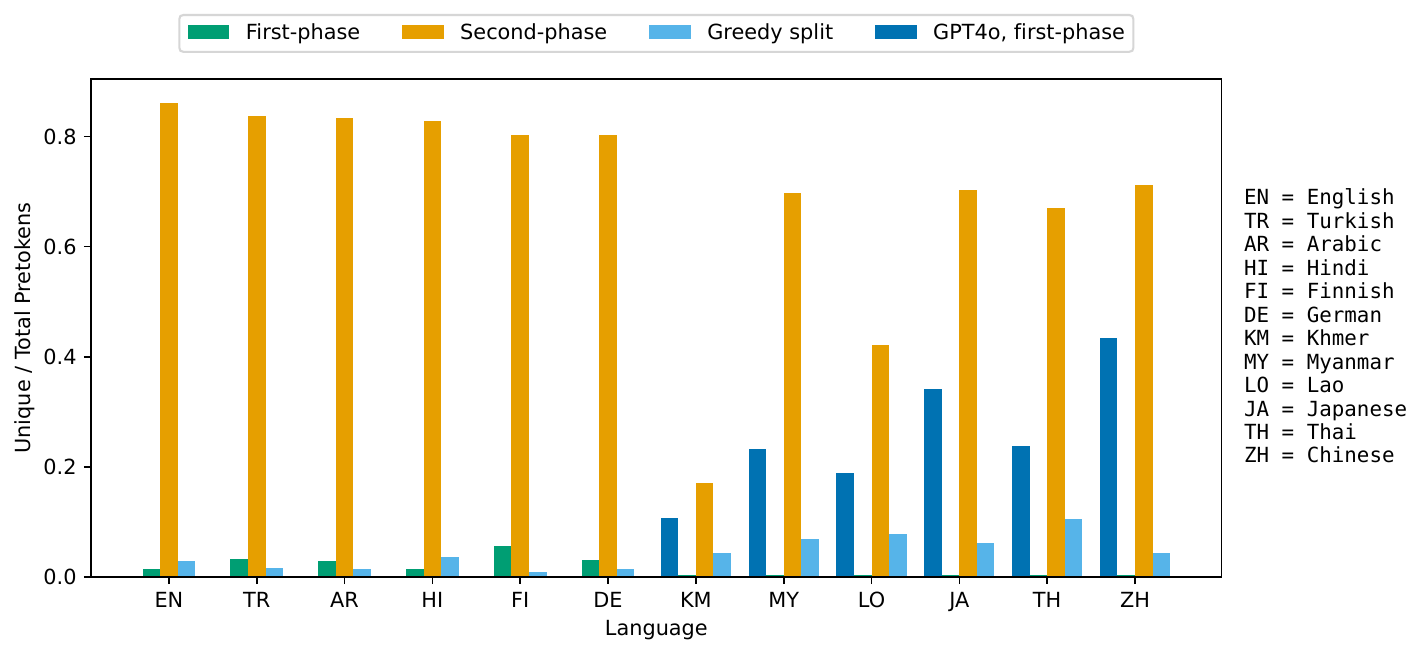}
    \caption{Type-token ratio (unique/total) by language for first-phase pretokens, second-phase supermerge candidates, and greedy-split candidates.
  Space-delimited scripts (EN\textrightarrow{}DE) present three calculations while non-space-delimited ones (KM\textrightarrow{}ZH) have an additional dark blue bar showing the first-phase ratio under the default GPT-4o regex. All results use 100K documents per language from CulturaX.}
    \label{fig:aggregationbylanguage}
\end{figure}

\cref{fig:aggregationbylanguage} shows the type-token ratio---the ratio of unique types to total tokens---computed separately for pretokens and for supermerge candidates. 
Results are shown for a range of languages using space-delimited scripts, as well as the six languages that require non-space-delimited script-specific pre-tokenization (see \cref{sec:aggregation} and \cref{app:pretokenization}). 
These results are computed over the first 100,000 documents in each language from the CulturaX multilingual dataset \citep{nguyen-etal-2024-culturax}. 
For the non-space-delimited cases, the dark blue bars show the first-phase ratio using the default GPT-4o regex, which is much higher than the first-phase ratio for the space-delimited ones (green bars).
Once the script-specific pre-tokenization is applied to the non-space-delimited text, the first-phase ratio drops to near zero, with the green bars hardly visible.
The second-phase supermerge candidate ratio (orange bars) is much higher than the first-phase ratio. 
There is some variation between languages depending on language structure and punctuation patterns. 

The two-phase BoundlessBPE process suggests a principled way to aggregate these further.
In the first phase we train ordinary BPE merge rules out to the desired vocabulary size $|V|$.
The counts of merges for each rule are non-increasing as they are added.
Let $c_{min}$ be the count of merges made by the last rule added.
A supermerge with a count less than $c_{min}$ will never be chosen in a vocabulary of size $|V|$, because the algorithm would prefer to have added ordinary merges with higher counts.
Thus we can use $c_{min}$ as a rigorous lower bound on the counts of possible supermerges that might be used.\footnote{For some non-space-delimited scripts there were not enough characters to fill the vocabulary, so in practice $\max(c_{min},15)$ was used. Note that we need the count of the last merge giving a vocabulary size $|V|$, so the first-phase SuperBPE run doesn't go far enough. It would be necessary to do a longer first-phase run for the full $|V|$ to find $c_{min}$ for SuperBPE.}

\cref{alg:ngrams} gives an algorithm to efficiently enumerate all $n$-grams of pretokens with at least an $n$-gram count of $c_{min}$ and a length from 2 to a maximum length of $L$.
While the performance depends strongly on the size of $c_{min}$, large enough values keep the number of $n$-grams that need to be counted to a reasonable number.
Without the bound, the number of $n$-grams to be counted quickly becomes prohibitively large for larger values of $L$.

To be as efficient as possible, we want to avoid counting any $n$-grams without the potential to reach $c_{min}$.
The key insight is that for an $n$-gram to have a count of $c_{min}$, the two $(n-1)$-grams contained within it must both have a count of at least $c_{min}$.
In the best case, every occurrence of an $(n-1)$-gram is also part of the $n$-gram, so the $n$-gram attains the same count of $c_{min}$; otherwise its count is lower.
The Apriori algorithm~\citep[][inter alia]{10.5555/645920.672836,10.1145/2452376.2452389} uses this form of pruning.
If we want all $n$-grams of size up to $L$, we generate $n$-grams in subsequent passes of increasing size $n = 1,\dots,L$.
We generate all $n$-grams of the current size, but only track the counts if both $(n-1)$-grams have a count of $c_{min}$.
After the pass counting size $n$, we filter those that didn't actually manage to have count $c_{min}$ at the end of the pass.

\begin{algorithm}[ht]
\begin{algorithmic}[1]
\Require Pretoken sequences with counts $S = \{(\textit{seq}, \textit{count})\}$
\Require Maximum $n$-gram length $L$
\Require Minimum count threshold $c_{min}$
\Ensure $n$-gram count map $C$ with all entries $\geq c_{min}$ and length $\geq 2$

\State $C \gets \{\}$ \Comment{Maps $n$-grams to counts; returns 0 for missing entries}

\For{$n = 1$ \textbf{to} $L$}
  \State $\textit{prev\_size} \gets |C|$

  \For{\textbf{each} $(\textit{seq}, \textit{count}) \in S$}
      \For{$i = 0$ \textbf{to} $|\textit{seq}| - n$}
          \If{$n > 1$ \textbf{and} $\min(C[\textit{seq}[i\!:\!i\!+\!n\!-\!1]],\; C[\textit{seq}[i\!+\!1\!:\!i\!+\!n]]) < c_{min}$}
              \State \textbf{continue} \Comment{Subgram bound}
          \EndIf
          \State $C[\textit{seq}[i\!:\!i\!+\!n]] \gets C[\textit{seq}[i\!:\!i\!+\!n]] + \textit{count}$
      \EndFor
  \EndFor

  \State Remove all $(g, c)$ from $C$ where $c < c_{min}$ \Comment{Prune}

  \If{$|C| = \textit{prev\_size}$} \Comment{No new $n$-grams at this length}
      \State \textbf{break}
  \EndIf
\EndFor

\State Remove all $(g, c)$ from $C$ where $|g| = 1$ \Comment{Drop unigrams}
\State \Return $C$
\end{algorithmic}
\caption{Frequent $n$-gram Counting with Subgram Pruning}
\label{alg:ngrams}
\end{algorithm}

We know that any supermerge not contained in one of these $n$-grams would not have a count high enough to be selected over a higher count ordinary merge.
However, these $n$-grams may overlap, double counting merges.
To avoid any overlap, we can partition the pretokens in a supermerge candidate in a left-to-right greedy manner.
Starting from the left of the run, match the longest possible $n$-gram, and then advance to the next position.
If no $n$-gram matches, advance to the next pretoken and repeat.
The counts of this greedy partitioning can be aggregated.
This gives the aggregation shown in the light blue bars of \cref{fig:aggregationbylanguage}, and we see that it has a low ratio across the range of languages. 

While this heuristic greedy splitting aggregates well, it does introduce an approximation to the BoundlessBPE process.
The splitting eliminates some potential merges that would have occurred over the split boundaries, so one might expect it to have fewer supermerges than normal BoundlessBPE. 

\begin{figure}[!ht]
   \centering
    \includegraphics[width=\linewidth]{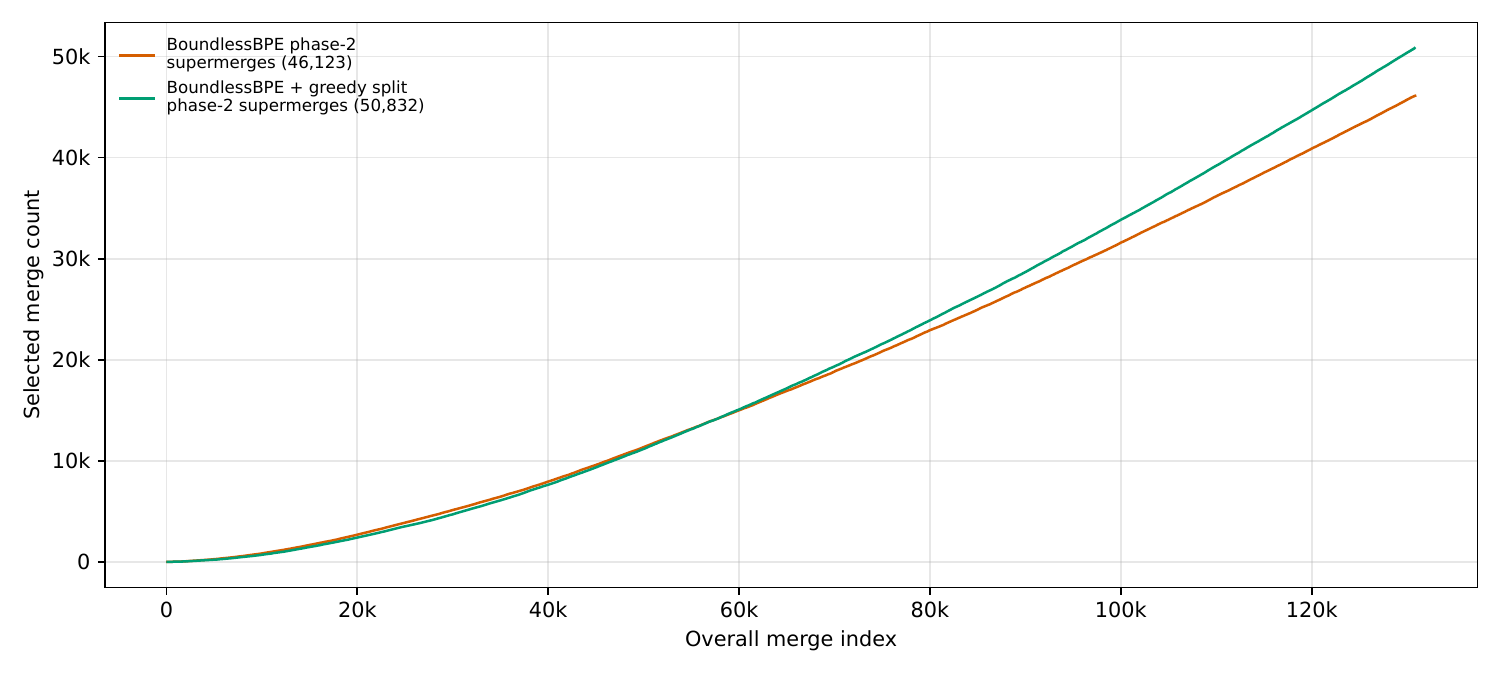}
    \caption{The merge counts of the selected supermerges for a BoundlessBPE run on 1M documents and a vocabulary size of 131,072, along with counts with the greedy splitting heuristic. The merge counts without the greedy split represent the number of merges that can actually be performed. Because of overestimated counts later in the merges, there were 50,832 supermerges selected with greedy splitting compared to 46,124 without.}
    \label{fig:greedysplit}
\end{figure}

\cref{fig:greedysplit} shows the counts of just the supermerges performed by a normal BoundlessBPE run, and one applied to the greedy split supermerge candidates.
Early in the process the supermerges have lower counts with the greedy splitting, but then it crosses over and the later merges have a higher count, so it ends up choosing more supermerges overall.
Consider a run of pretokens \texttt{[..., X, A, B, Y, ...]} as a potential supermerge candidate, followed by a greedy split producing \texttt{[..., X], [A, B], [Y, ...]}.
The number of \texttt{[A,B]} merges is the same as before.
However, suppose there is an \texttt{[X,A]} merge or a \texttt{[B,Y]} merge that happens before \texttt{[A,B]}.
Without the greedy split, it would decrease the pairwise count of \texttt{[A,B]} merges, since some of them are now absorbed into the earlier merge.
Here this is avoided, because the greedy split isolated the \texttt{[A,B]}.
Consequently, we have eliminated some potential merges through the splits, and pairwise counts will be lower than without.
However, over time, the estimates of the potential supermerge counts will be a slight overestimate, leading to more supermerges being chosen with higher counts than would be actually possible in inference.
The merge counts could be made correct by also doing greedy left-to-right splitting at inference time.
Then supermerge inference would match training, with decisions based on correct counts, but subject to a heuristic.
It isn't obvious which would perform better in practice, and further research is warranted on this approach.

\section{Estimation of Original BoundlessBPE Training Times}
\label{app:extrapolation}

The original BoundlessBPE implementation saves checkpoints every 8192 iterations, so a variety of vocabulary sizes can be examined in one run; however it is extremely slow.
Runs for 10K, 20K, and 50K documents ran to the full vocabulary size 131,072. We would like to estimate the runtime for 100K and 200K runs, using the vocabulary sizes that finished.
A simple linear fit of time vs. vocabulary size has an $R^2 > 0.99$ for all series.
We used the extrapolated time for these two points in \cref{fig:timing}.
The original implementation does not index the pretokens containing potential supermerges, so it does not have the sublinear performance seen in \cref{fig:timingvsvocab}.

\begin{figure}[!ht]
    \centering
    \includegraphics[width=\linewidth]{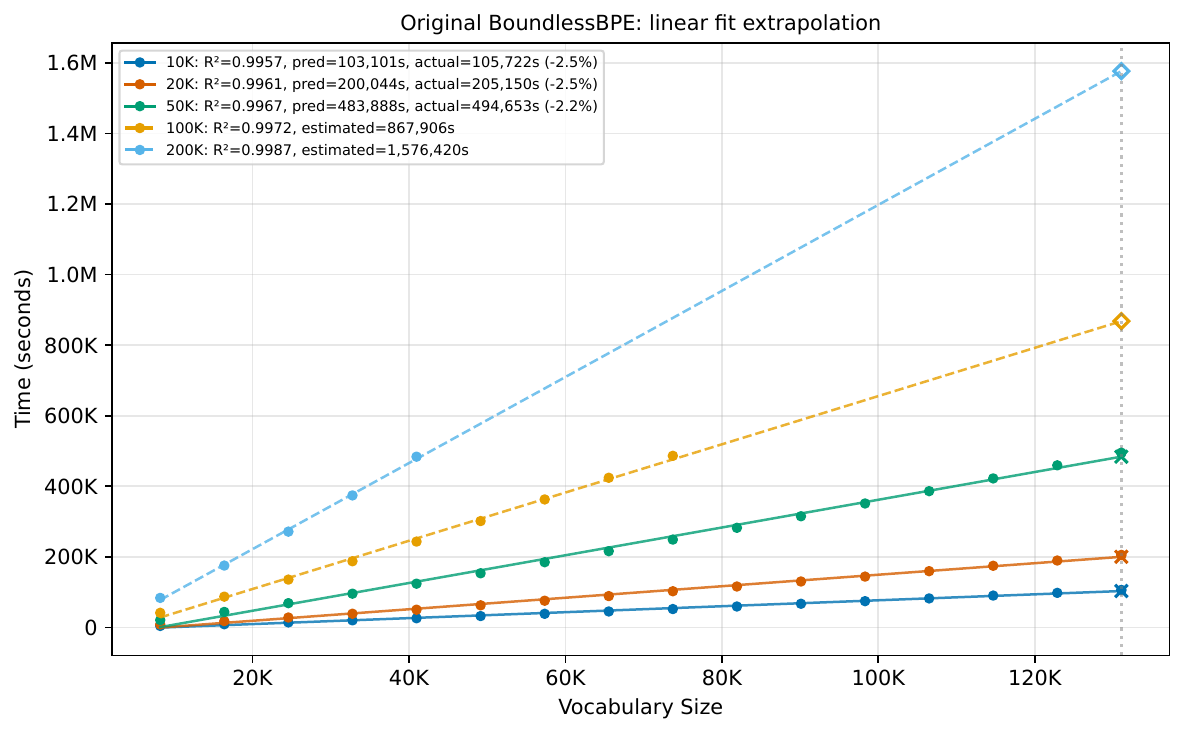}
    \caption{Time vs. vocabulary size using the original BoundlessBPE implementation, for various numbers of documents. A linear fit of the data works well, and allows an extrapolation of the final time for 100K and 200K documents.}
    \label{fig:extrapolation}
\end{figure}

\section{Detailed Training Timing}

\cref{tab:timing-combined} compares mean training times (over 3 runs) for the Python and Rust implementations, with a vocabulary size of 131,072.
Times are in seconds, with the Rust speedup shown as $\times$.
The table is organized into four sections corresponding to the stages of two pipelines.
The BoundlessBPE pipeline combines the first two sections: a standard BPE run (which serves as the first phase) and the BoundlessBPE second phase.
The SuperBPE pipeline combines the third and fourth sections: a BPE run matched to produce the same number of ordinary merges as the BoundlessBPE run, followed by the SuperBPE second phase with the same number of supermerges.
This ensures both pipelines produce vocabularies of the same size and composition, making their training times directly comparable.

In BPE, a majority of time is spent in the pre-tokenization phase.
In particular, much of the time is spent in the regex libraries.
The Python library \texttt{regex}\footnote{\scriptsize{\url{https://github.com/mrabarnett/mrab-regex}}} is very mature and written in C, so Rust has no great advantage for time spent within the library, giving the small speedups seen.
Both the BoundlessBPE and SuperBPE second-phase runs spend more time in the merge process, which receives a substantial speedup in moving this procedural code to Rust.
Note that these implementations are currently single-threaded; additional speedups could result from adding parallelism through multi-threading.

\begin{table*}[tbh]
\centering
\small
\begin{tabular}{ll rrr rrr rrr}
\toprule
 & & \multicolumn{3}{c}{Pre-tokenization} & \multicolumn{3}{c}{Merge} & \multicolumn{3}{c}{Total} \\
\cmidrule(lr){3-5} \cmidrule(lr){6-8} \cmidrule(lr){9-11}
Method & Docs & Py & Rust & $\times$ & Py & Rust & $\times$ & Py & Rust & $\times$ \\
\midrule
\multirow{7}{*}{BPE} & 10K & 11 & 6 & 1.7$\times$ & 43 & 8 & 5.0$\times$ & 62 & 18 & 3.4$\times$ \\
 & 20K & 23 & 13 & 1.8$\times$ & 62 & 14 & 4.3$\times$ & 94 & 31 & 3.0$\times$ \\
 & 50K & 59 & 35 & 1.7$\times$ & 104 & 26 & 4.1$\times$ & 173 & 65 & 2.7$\times$ \\
 & 100K & 118 & 69 & 1.7$\times$ & 154 & 40 & 3.8$\times$ & 284 & 115 & 2.5$\times$ \\
 & 200K & 234 & 141 & 1.7$\times$ & 235 & 65 & 3.6$\times$ & 484 & 212 & 2.3$\times$ \\
 & 500K & 582 & 354 & 1.6$\times$ & 412 & 125 & 3.3$\times$ & 1,015 & 491 & 2.1$\times$ \\
 & 1M & 1,151 & 706 & 1.6$\times$ & 636 & 207 & 3.1$\times$ & 1,815 & 930 & 2.0$\times$ \\
\midrule
\multirow{7}{*}{\shortstack[l]{BoundlessBPE\\second pass}} & 10K & 17 & 7 & 2.5$\times$ & 113 & 11 & 10.1$\times$ & 184 & 31 & 6.0$\times$ \\
 & 20K & 38 & 15 & 2.5$\times$ & 217 & 22 & 9.8$\times$ & 321 & 57 & 5.7$\times$ \\
 & 50K & 116 & 41 & 2.8$\times$ & 485 & 55 & 8.8$\times$ & 701 & 138 & 5.1$\times$ \\
 & 100K & 240 & 84 & 2.9$\times$ & 908 & 108 & 8.4$\times$ & 1,333 & 271 & 4.9$\times$ \\
 & 200K & 505 & 172 & 2.9$\times$ & 1,656 & 209 & 7.9$\times$ & 2,448 & 534 & 4.6$\times$ \\
 & 500K & 1,301 & 431 & 3.0$\times$ & 3,509 & 502 & 7.0$\times$ & 5,388 & 1,297 & 4.2$\times$ \\
 & 1M & 2,591 & 844 & 3.1$\times$ & 6,253 & 944 & 6.6$\times$ & 9,868 & 2,510 & 3.9$\times$ \\
\midrule
\multirow{7}{*}{\shortstack[l]{BPE (matched\\for SuperBPE)}} & 10K & 12 & 6 & 1.8$\times$ & 39 & 8 & 4.9$\times$ & 53 & 15 & 3.5$\times$ \\
 & 20K & 23 & 14 & 1.7$\times$ & 53 & 13 & 3.9$\times$ & 78 & 28 & 2.8$\times$ \\
 & 50K & 60 & 35 & 1.7$\times$ & 94 & 25 & 3.8$\times$ & 157 & 62 & 2.5$\times$ \\
 & 100K & 118 & 71 & 1.6$\times$ & 140 & 39 & 3.6$\times$ & 262 & 113 & 2.3$\times$ \\
 & 200K & 235 & 143 & 1.6$\times$ & 210 & 63 & 3.3$\times$ & 452 & 211 & 2.1$\times$ \\
 & 500K & 585 & 360 & 1.6$\times$ & 372 & 121 & 3.1$\times$ & 970 & 490 & 2.0$\times$ \\
 & 1M & 1,160 & 717 & 1.6$\times$ & 581 & 215 & 2.7$\times$ & 1,761 & 946 & 1.9$\times$ \\
\midrule
\multirow{7}{*}{\shortstack[l]{SuperBPE\\second pass}} & 10K & 20 & 7 & 2.7$\times$ & 106 & 11 & 10.1$\times$ & 144 & 24 & 5.9$\times$ \\
 & 20K & 43 & 16 & 2.7$\times$ & 210 & 21 & 9.9$\times$ & 286 & 50 & 5.7$\times$ \\
 & 50K & 115 & 42 & 2.8$\times$ & 476 & 53 & 9.0$\times$ & 667 & 128 & 5.2$\times$ \\
 & 100K & 240 & 84 & 2.9$\times$ & 875 & 103 & 8.5$\times$ & 1,261 & 256 & 4.9$\times$ \\
 & 200K & 502 & 171 & 2.9$\times$ & 1,617 & 200 & 8.1$\times$ & 2,372 & 508 & 4.7$\times$ \\
 & 500K & 1,289 & 429 & 3.0$\times$ & 3,309 & 480 & 6.9$\times$ & 5,204 & 1,250 & 4.2$\times$ \\
 & 1M & 2,559 & 843 & 3.0$\times$ & 5,709 & 904 & 6.3$\times$ & 9,374 & 2,426 & 3.9$\times$ \\
\bottomrule
\end{tabular}
\caption{Training time comparison: Python vs.\ Rust implementation (seconds), averaged over 3 runs.
Total includes all components; pre-tokenization and merge are shown separately.}
\label{tab:timing-combined}
\end{table*}

\end{document}